\documentclass[11pt, a4paper, onecolumn, thu]{thuc3i}

\usepackage[authoryear, sort&compress, square]{natbib}
\usepackage{fontawesome5}

\usepackage[utf8]{inputenc} 
\usepackage[T1]{fontenc}    
\usepackage{hyperref}       
\usepackage{url}            
\usepackage{booktabs}       
\usepackage{amsfonts}       
\usepackage{nicefrac}       
\usepackage{microtype}      
\usepackage{xcolor}         

\usepackage{amsmath}   
\usepackage{amsthm}    

\usepackage{enumitem}  

\usepackage{bbm}

\definecolor{lightblue}{RGB}{220,235,250}
\usepackage[most,skins,theorems]{tcolorbox}
\tcbset{
  takeawaysbox/.style={
    title=Takeaways,
    colback=lightblue!80,
    colframe=black,
    fonttitle=\bfseries\small,
    coltitle=white,
    colbacktitle=black,
    enhanced,
    attach boxed title to top left={xshift=2.5mm,yshift=-2.5mm},
    boxed title style={rounded corners, size=small, colframe=black, colback=black},
    width=\linewidth,
    arc=3.5mm
  }
}

\usepackage{multirow, colortbl}
\usepackage{graphicx}
\definecolor{lightblue}{RGB}{220,235,250}

\usepackage{wrapfig}

\newcommand{\tableDashLine}[1]{%
  \arrayrulecolor{gray!55}\specialrule{0.4pt}{1pt}{1pt}\arrayrulecolor{black}%
}

\setheadertext{ZEDA}

\title{Post-Trained MoE Can Skip Half Experts via Self-Distillation}

\reportnumber{} 

\renewcommand{\today}{2026-05-19}

\author[*1]{Xingtai Lv}
\author[*1]{Li Sheng}
\author[*1,5]{Kaiyan Zhang}
\author[1]{Yichen You}
\author[1,4]{Siyan Gao}
\author[1]{Xueheng Luo}
\author[1]{Yuxin Zuo}
\author[1]{Yuchen Fan}
\author[1,5]{Junlin Yang}
\author[2]{Ganqu Cui}
\author[3]{Bingning Wang}
\author[4]{Fan Yang}
\author[$\dagger$1,2]{Youbang Sun}
\author[$\dagger$1,2]{Ning Ding}
\author[$\dagger$1,2]{Bowen Zhou}

\affil[1]{\thepa{}{}}
\affil[2]{Shanghai AI Lab}
\affil[3]{WeChat AI}
\affil[4]{Kuaishou Technology}
\affil[5]{Frontis.AI}

\role[*]{Equal Contributions}
\role[\dagger]{Corresponding Authors}

\resource{\faEnvelope[regular]}{lvxt24@mails.tsinghua.edu.cn}
\resource{\faGithub}{\href{https://github.com/TsinghuaC3I/ZEDA}{TsinghuaC3I/ZEDA}}

\begin{abstract}
Mixture-of-Experts (MoE) scales language models efficiently through sparse expert activation, and its dynamic variant further reduces computation by adjusting the activated experts in an input-dependent manner.
Existing dynamic MoE methods usually rely on pre-training from scratch or task-specific adaptation, leaving the practical conversion of fully trained MoE underexplored. 
Enabling such adaptation would directly alleviate the inference costs by allowing easy tokens to bypass unnecessary expert during serving.
This paper introduces \textbf{Z}ero-\textbf{E}xpert Self-\textbf{D}istillation \textbf{A}daptation (\textbf{ZEDA}), a low-cost framework that transforms post-trained static MoE models into efficient dynamic ones. To stabilize this architectural conversion, ZEDA injects parameter-free zero-output experts into each MoE layer and adapts the augmented model through two-stage self-distillation, utilizing the original MoE as a frozen teacher and applying a group-level balancing loss.
On Qwen3-30B-A3B and GLM-4.7-Flash across 11 benchmarks spanning math, code, and instruction following, ZEDA eliminates over 50\% of expert FLOPs at marginal accuracy loss.
It outperforms the strongest dynamic MoE baseline by 6.1 and 4.0 points on the two models, and delivers \textasciitilde1.20× end-to-end inference speedup.

\end{abstract}

\begin{document}

\maketitle


\begin{figure}[h]
  \centering
  \includegraphics[width=\textwidth]{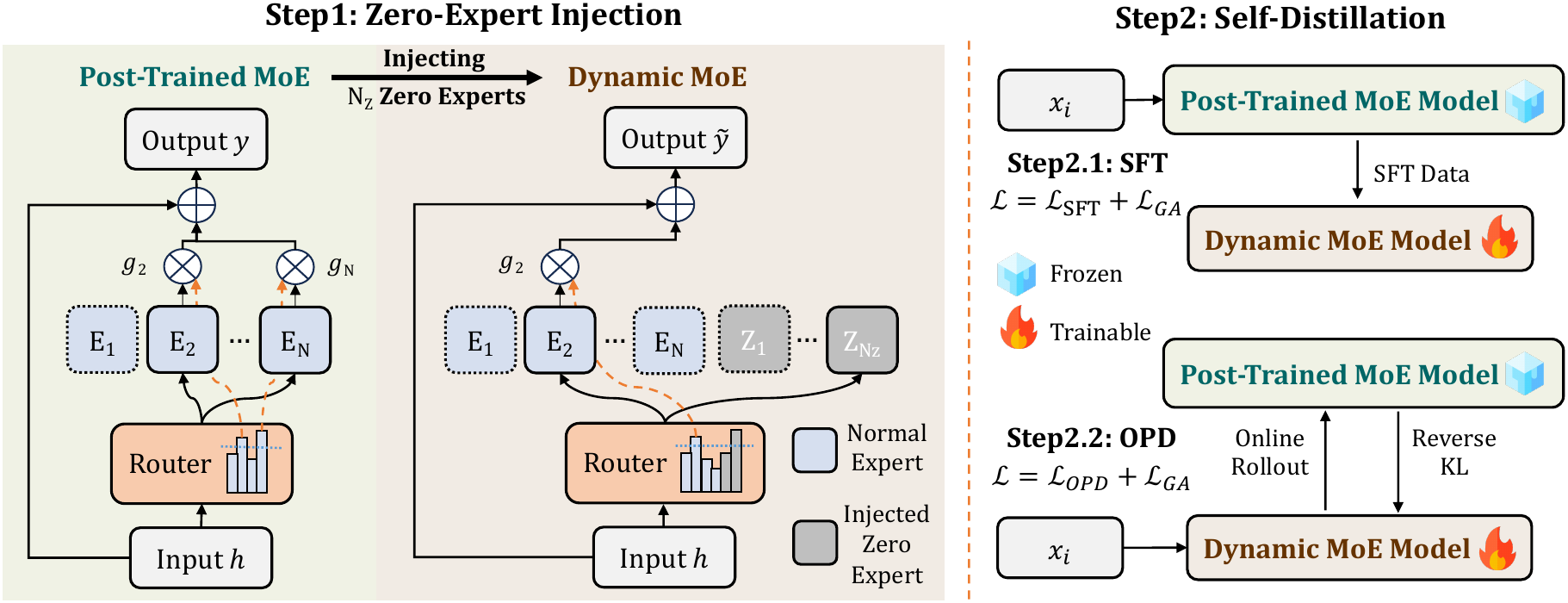}
  \caption{Illustration of ZEDA. ZEDA leverages the post-trained MoE to initialize the dynamic MoE (with zero-expert injection) and further utilizes it as a teacher model for distillation.}
  \label{fig:zesa}
\end{figure}

\newpage
\begingroup
\setlength{\baselineskip}{0.85\baselineskip}
\tableofcontents
\endgroup
\newpage

\section{Introduction}
\label{sec:intro}

Mixture-of-Experts (MoE) architectures have significantly advanced the scaling of large language models (LLMs) by increasing model capacity while keeping bounded per-token computation~\citep{lepikhin2020gshard,fedus2022switch,du2022glam, dai2024deepseekmoe, jiang2024mixtral}. Building upon this foundation, a variant we refer to as \emph{dynamic MoE}, further introduces token-level dynamism that adjusts the number of activated experts, enabling an input-dependent allocation of computation budgets~\citep{jin2024moe++, team2025longcat, wu2025grove, guo2024dynamic, chaudhari2026moe, zeng2024adamoe}. Many studies have demonstrated that easy tokens can be processed with substantially fewer experts without compromising output quality, making dynamic MoE a principled route to inference-time efficiency~\citep{lu2024not, jin2024moe++, team2025longcat, zeng2024adamoe, huang2024harder}.

Most existing approaches to dynamic MoE concentrate on either pre-training dynamic MoE models from scratch~\citep{jin2024moe++,team2025longcat, chaudhari2026moe} or adapting a pre-trained base model into a task-specific dynamic MoE~\citep{zeng2024adamoe}, leaving the migration of fully trained MoE models largely unexplored. Yet in practical deployment, MoE models have typically undergone an extensive training pipeline encompassing both pre-training and post-training such as supervised fine-tuning (SFT), reinforcement learning (RL), and on-policy distillation (OPD)~\citep{qwen35blog, zeng2026glm5, deepseekai2026deepseekv4}. We refer to such models as \emph{post-trained MoE} throughout this paper. If such a post-trained static MoE model could be converted into a more efficient dynamic counterpart with the architecture and primary training already finalized, the resulting inference savings would be of tremendous practical value given the ever-growing serving costs and demand.

However, directly applying existing dynamic MoE methods to such models risks disrupting the carefully calibrated routing and capability distributions established during the full training pipeline. In this paper, we focus on exploring \emph{whether a post-trained MoE model can be cost-effectively migrated into a more efficient dynamic MoE without sacrificing its established capabilities}.

We introduce \textbf{Z}ero-\textbf{E}xpert Self-\textbf{D}istillation \textbf{A}daptation (\textbf{ZEDA}), transforming a post-trained MoE model into a dynamic one with faster inference at minimal adaptation cost. 
ZEDA injects parameterless \emph{zero experts}~\citep{jin2024moe++,team2025longcat}, whose outputs are identically zero, into the existing expert pool of a post-trained MoE model. This expands the router candidate pool with zero-computation experts while the activation number remains unchanged, naturally reducing active normal experts.
The augmented model is then adapted through a two-stage self-distillation process comprising SFT~\citep{ouyang2022training} and OPD~\citep{gu2023minillm, agarwal2024policy, lu2025onpolicydistillation}, using the original MoE as a fixed teacher, to recover performance under the new dynamic routing regime.
To make this architectural conversion stable, ZEDA further introduces a Group Auxiliary Loss $\mathcal{L}_{GA}$ that regulates the relative activation frequency between normal experts and zero experts while preserving the learned routing structures among normal experts.

Experiments on Qwen3-30B-A3B~\citep{yang2025qwen3} and GLM-4.7-Flash~\citep{zeng2025glm} across 11 benchmarks spanning math, code, and instruction following demonstrate the effectiveness of ZEDA. Our method successfully migrates post-trained MoE models into dynamic ones in less than 31 hours for Qwen and 62 hours for GLM on 8 NVIDIA H200 GPUs. This adaptation eliminates over half of the expert computation and achieves an inference speedup around 20\%, while incurring only a marginal accuracy loss compared with the original model.
ZEDA outperforms the strongest baseline by an average of 6.1 points on Qwen and 4.0 points on GLM, and also achieves the best overall performance among our proposed variants.
Through detailed illustrative visualizations and analysis, the dynamic characteristics of the zero expert activation and the operating mechanisms of ZEDA are clearly revealed. 
The following are several key takeaways:


\begin{tcolorbox}[takeawaysbox]
\begin{enumerate}[leftmargin=0.2em]
    \item ZEDA cost-effectively converts post-trained MoE models into dynamic ones across diverse domains, reducing over half of the MoE computation while maintaining performance~\textbf{($\S$~\ref{experiments})}.

    \item Injecting zero experts and group-level balancing strategy minimize disruptions to original routing distribution, facilitating stable adaptation of post-trained MoE~\textbf{($\S$~\ref{sec:method})}.
    
    \item Zero-expert activation (compute allocation) intrinsically correlates with the teacher-student distribution gap, model uncertainty, and response patterns, not overall task difficulty~\textbf{($\S$~\ref{sec:analysis_rze})}.
\end{enumerate}
\end{tcolorbox}
\section{Method}
\label{sec:method}

We propose \textbf{Z}ero-\textbf{E}xpert Self-\textbf{D}istillation \textbf{A}daptation (\textbf{ZEDA}), a method that transforms a post-trained MoE model into a dynamic one with faster inference at minimal adaptation cost, by augmenting each MoE module with zero experts and adapting the expanded model through self-distillation.
In the following, we present the overall adaptation framework in Section~\ref{sec:framework}, and then introduce the Group Auxiliary Loss $\mathcal{L}_{GA}$ that regulates zero expert utilization in Section~\ref{sec:gal}.

\subsection{Adaptation Framework}
\label{sec:framework}

ZEDA first injects zero experts~\citep{jin2024moe++}, whose outputs are identically zero, into a post-trained MoE, architecturally converting it into a dynamic one whose activated normal experts number varies across tokens. The augmented model is then adapted through the two-stage self-distillation with the original post-trained MoE as a fixed teacher, yielding a more efficient dynamic MoE with negligible performance loss.

\paragraph{Zero-Expert Injection.}
Consider a post-trained MoE model where each MoE module contains $N$ normal experts $\mathcal{E} = \{E_1, \dots, E_N\}$ and activates $K$ of them per token. For an input hidden state $h$, the router selects a top-$K$ subset $\mathcal{S}(h) \subseteq \mathcal{E}$ and produces
$y(h) = \sum_{i \in \mathcal{S}(h)} g_i(h)\, E_i(h)$,
where $g_i(h)$ is the normalized routing weight for expert $E_i$.

ZEDA introduces $N_Z$ additional experts $\mathcal{Z} = \{Z_1, \dots, Z_{N_Z}\}$ that satisfy $Z_j(h) = 0$ for all $j$, referred to as \emph{zero experts}. The augmented expert pool $\mathcal{E}' = \mathcal{E} \cup \mathcal{Z}$ expands the router from $N$ to $N + N_Z$ candidates while the top-$K$ budget remains unchanged. The dynamic MoE output becomes
\begin{equation}
\label{eq:dynamic_moe}
\tilde{y}(h) = \sum_{i \in \tilde{\mathcal{S}}(h) \cap \mathcal{E}} \tilde{g}_i(h)\, E_i(h),
\end{equation}
where $\tilde{\mathcal{S}}(h)$ denotes the top-$K$ set selected from $\mathcal{E}'$ and $\tilde{g}_i(h)$ is the corresponding routing weight. 
Because zero experts contribute no computation, selecting them reduces the number of active normal experts, yielding token-dependent computation without modifying the normal expert parameters.
We also compare the zero expert with another zero-computation alternative, \emph{copy expert}, which outputs its input, in Appendix~\ref{sec:zero_vs_copy}, showing that copy experts induce both scale and direction mismatches.

For router initialization, the original router parameters for the $N$ normal experts are kept unchanged. The new parameters for the $N_Z$ zero experts are drawn from a Gaussian distribution matching the mean and variance of the original router parameters in the same module, preserving the post-trained scale of router logits while inserting new routing options.

\paragraph{Two-Stage Self-Distillation.}
ZEDA then adapts the augmented model via self-distillation, using the original MoE as a fixed teacher. The adaptation proceeds in two stages, supervised fine-tuning (SFT) followed by on-policy distillation (OPD). Let $\pi_T$ denote the teacher (original MoE) distribution, $\pi_\theta$ the student (augmented model) distribution, and $\mathcal{D}$ the prompt set used for adaptation.

\begin{itemize}[leftmargin=1em]
\item The SFT stage trains $\pi_\theta$ on responses sampled from the teacher $\pi_T$. The training loss is:
\begin{equation}
\label{eq:sft}
\mathcal{L} = \mathcal{L}_{\mathrm{SFT}}+ \mathcal{L}_{\mathrm{GA}} = -\mathbb{E}_{x \sim \mathcal{D},\, y \sim \pi_T(\cdot \mid x)} \left[ \sum_{t=1}^{|y|} \log \pi_\theta(y_t \mid x, y_{<t}) \right] + \mathcal{L}_{\mathrm{GA}},
\end{equation}
where $x$ is a prompt from $\mathcal{D}$, $y = (y_1, \ldots, y_{|y|})$ is a teacher-sampled response, and $\mathcal{L}_{\mathrm{GA}}$ is the group auxiliary loss introduced in Section~\ref{sec:gal}.

\item The subsequent OPD stage~\citep{gu2023minillm,agarwal2024policy} shifts to on-policy learning, where responses are sampled from the current student $\pi_\theta$ and the teacher evaluates the same trajectories to supply token-level targets. Following Thinking Machines \citep{lu2025onpolicydistillation}, we cast the sampled-token reverse KL objective as a reward signal and optimize it within the policy optimization framework, yielding the training loss:
\begin{equation}
\label{eq:opd_loss}
\mathcal{L} = \mathcal{L}_{\mathrm{OPD}}+ \mathcal{L}_{\mathrm{GA}} = \mathbb{E}_{x \sim \mathcal{D},\, y \sim \pi_\theta(\cdot \mid x)} \left[ \sum_{t=1}^{|y|} \mathrm{KL}\!\left( \pi_\theta(\cdot \mid x, y_{<t}) \,\|\, \pi_T(\cdot \mid x, y_{<t}) \right) \right] + \mathcal{L}_{\mathrm{GA}}.
\end{equation}
\end{itemize}

The SFT stage stabilizes the initial transition from a static to a dynamic MoE, and the OPD stage further aligns the student with the teacher under the student's own rollout distribution.


\subsection{Group Auxiliary Loss}
\label{sec:gal}

ZEDA incorporate the Group Auxiliary Loss $\mathcal{L}_{GA}$ to regulate the relative activation frequency between normal experts and zero experts, thereby controlling the zero expert activation ratio $r_{ZE}$.

\paragraph{Auxiliary Loss.}
$\mathcal{L}_{GA}$ is derived from the vanilla auxiliary load balancing loss $\mathcal{L}_{A}$~\citep{lepikhin2020gshard,fedus2022switch}, which encourages uniform routing across all experts. $\mathcal{L}_{A}$ is defined through the batch $\mathcal{B}$:
\begin{equation}
    \mathcal{L}_{A} = \alpha \cdot \frac{N + N_Z}{K} \cdot \sum_{i=1}^{N+N_Z} f_i \cdot P_i, \quad
     \text{where} \quad f_i = \frac{1}{|\mathcal{B}|} \sum_{h \in \mathcal{B}} \mathbbm{1} \!\left\{ i \in \tilde{\mathcal{S}}(h) \right\},
    P_i = \frac{1}{|\mathcal{B}|} \sum_{h \in \mathcal{B}} \tilde{g}_i(h).
    \label{eq:aux}
\end{equation}
Here, $f_i$ denotes the fraction of tokens in a batch $\mathcal{B}$ routed to expert $i$, $P_i$ is the mean routing probability assigned to expert $i$ over $\mathcal{B}$, and $\alpha$ is a scalar loss coefficient.
However, applying $\mathcal{L}_A$ directly in ZEDA is problematic. A post-trained MoE model exhibits non-uniform, input-dependent routing patterns over normal experts, and enforcing expert-level uniformity would disrupt these learned distributions, degrading model performance. Appendix~\ref{app:aux-loss-comparison} presents a dedicated experiment comparing $\mathcal{L}_A$ and $\mathcal{L}_{GA}$.

\paragraph{Group Load Balancing Loss.}
The objective of ZEDA is to regulate zero expert utilization while preserving the relative routing structure among normal experts. This motivates a group-level balancing strategy in which the $N$ normal experts form a group $\mathcal{E}$ and the $N_Z$ zero experts form a group $\mathcal{Z}$, with balancing applied only between the two groups. The Group Auxiliary Loss is defined as
\begin{equation}
    \mathcal{L}_{GA} = \alpha \cdot \frac{N + N_Z \cdot w}{K} \cdot \left( \frac{f_{\mathcal{E}} \cdot P_{\mathcal{E}}}{N} + \frac{f_{\mathcal{Z}} \cdot P_{\mathcal{Z}}}{N_Z \cdot w} \right),
    \label{eq:gal}
\end{equation}
\begin{equation}
\text{where} \quad f_{\mathcal{E}} = \sum_{i \in \mathcal{E}} f_i, \quad P_{\mathcal{E}} = \sum_{i \in \mathcal{E}} P_i, \quad f_{\mathcal{Z}} = \sum_{i \in \mathcal{Z}} f_i, \quad P_{\mathcal{Z}} = \sum_{i \in \mathcal{Z}} P_i.
\end{equation}
$w > 0$ is the relative weight of the zero-expert group, and a larger $w$ encourages higher $r_{ZE}$.


Analogously to $\mathcal{L}_A$, minimizing $\mathcal{L}_{GA}$ drives the two groups toward an equilibrium in which the expected number of activated normal experts $K_{\mathcal{E}}$ and zero experts $K_{\mathcal{Z}}$ satisfy $K_{\mathcal{E}} : K_{\mathcal{Z}} = N : N_Z \cdot w$, yielding a target $r_{ZE} = (N_Z \cdot w)/(N + N_Z \cdot w).$
Since the constraint is imposed only at the group level, it does not explicitly flatten the routing distribution within the normal-expert group, which makes it better aligned with post-trained MoE adaptation.
$\mathcal{L}_{GA}$ drives $r_{ZE}$ toward the target value, while the other loss component ($\mathcal{L}_{\mathrm{SFT}}$ or $\mathcal{L}_{\mathrm{OPD}}$) optimizes performance. Under the joint influence, the model reaches a trade-off, causing $r_{ZE}$ to converge to an appropriate value.

\section{Experiments}
\label{experiments}

\subsection{Experimental Setup}
\label{sec:settings}

\paragraph{Models.}
To evaluate the generalizability of ZEDA across different backbone architectures, two post-trained MoE models are selected: Qwen3-30B-A3B~\citep{yang2025qwen3} and GLM-4.7-Flash~\citep{zeng2025glm}. Qwen3-30B-A3B is consistently used in Thinking mode throughout all experiments. The two models differ in scale and expert configuration. Qwen3-30B-A3B contains $N{=}128$ normal experts with $K{=}8$ activated per token, while GLM-4.7-Flash has $N{=}64$ and $K{=}4$. Following LongCat~\citep{team2025longcat}, the number of injected zero experts $N_Z$ is set to 64 and 32 for Qwen3-30B-A3B and GLM-4.7-Flash, respectively.

\paragraph{Evaluation Setup.}
To comprehensively assess the post-adaptation performance, ZEDA is evaluated on 11 benchmarks spanning 3 categories. For math reasoning, the benchmarks include AIME 24, AIME 25, AIME 26~\citep{li2024numinamath}, GSM8K~\citep{cobbe2021training}, and MATH-500~\citep{lightman2023let}. For code generation, the benchmarks include LiveCodeBench v5 (LCB v5), LiveCodeBench v6 (LCB v6)~\citep{jain2024livecodebench}, HumanEval+~\citep{liu2023your}, and MBPP+~\citep{liu2023your}. HumanEval+ and MBPP+ are two code generation benchmarks introduced by EvalPlus~\citep{liu2023your}. For instruction following, the benchmarks include IFEval~\citep{zhou2023instruction} and IFBench~\citep{pyatkin2025generalizing}. 
All evaluations adopt a temperature of 0.6, a top-$p$ value of 0.95, and a top-$k$ value of 20, with a maximum generation length of 38k tokens following the Qwen3 setting~\citep{yang2025qwen3}. 
We report avg@32 for AIME24, AIME25, and AIME26 to reduce variance on these small-scale competition benchmarks, avg@8 for the 4 coding benchmarks, and avg@1 for all remaining benchmarks. 
Following the conventions of Qwen3~\citep{yang2025qwen3} and IFBench~\citep{pyatkin2025generalizing}, results on IFEval and IFBench are reported as strict prompt accuracy and loose prompt accuracy, respectively.

\paragraph{Implementation Details.}
For the inference efficiency of the adapted dynamic MoE, the relative weight $w$ in $\mathcal{L}_{GA}$ (Eq.~\ref{eq:gal}) is set to 2, which drives the target $r_{ZE}$ toward 50\%, and the loss coefficient $\alpha$ is set to 0.1. Ablation studies on $w$ and $\alpha$ are presented in Section~\ref{sec:analyze_effect_of_weight} and Section~\ref{sec:ablation}, respectively. The self-distillation data consists of 60k prompts in total. It consists of 17k math prompts and 15k coding prompts randomly sampled from NVIDIA AceReason-1.1-SFT~\citep{liu2025acereason}, together with 28k chat prompts randomly sampled from NVIDIA Llama-Nemotron-Post-Training-Dataset~\citep{bercovich2025llama}.
In the SFT stage, the learning rate is set to $2 \times 10^{-5}$. The subsequent stage employs Sampled-Token OPD with a learning rate of $5 \times 10^{-6}$ for Qwen3-30B-A3B and $1 \times 10^{-6}$ for GLM-4.7-Flash, a batch size of 16 prompts $\times$ 2 sampled responses, a sampling temperature of 1.0, a maximum generation length of 32k tokens, and runs for 320 training steps. All experiments are conducted on the slime~\citep{slime_github}, SGLang~\citep{zheng2024sglang}, and Megatron~\citep{shoeybi2019megatron} codebases, and on NVIDIA H200 and H20 GPUs.

\paragraph{Baselines.}
AdaMoE~\citep{zeng2024adamoe} and the Dynamic Skipping method in~\citep{lu2024not} serve as the dynamic routing baselines. 
We further propose three variants to evaluate the efficacy of ZEDA’s components. ZEDA$_{\mathrm{SFT}}$, which applies only the SFT stage of ZEDA, is included to isolate the contribution of OPD. To validate the dynamic expert selection mechanism, we propose Naive Expert Truncation (NET), a straightforward variant of ZEDA that directly halves the number of activated experts in the original MoE model. NET is combined with SFT alone or SFT followed by OPD, yielding NET$_{\mathrm{SFT}}$ and NET$_{\mathrm{SFT \to OPD}}$, respectively.

\subsection{Main Results}

\paragraph{Performance.}
Table~\ref{tab:main-results} summarizes the performance of all methods on 11 benchmarks spanning mathematical reasoning, code generation, and instruction following. Compared with the original post-trained MoE, ZEDA incurs only a marginal average accuracy loss while eliminating over half of the expert computation, and even surpasses the original model on several individual benchmarks such as IFBench, demonstrating the practical utility of the dynamic MoE models produced by ZEDA. Among all baselines, ZEDA achieves the highest average evaluation scores on both Qwen3-30B-A3B and GLM-4.7-Flash, indicating its effectiveness and robustness across architectures.
Furthermore, ZEDA achieves superior overall performance over all three variants, ZEDA$_{\mathrm{SFT}}$, NET$_{\mathrm{SFT}}$ and NET$_{\mathrm{SFT \to OPD}}$, demonstrating the contributions of OPD and the dynamic expert selection mechanism.
Moreover, the dynamic routing baselines exhibit severe capability imbalances, where AdaMoE collapses on hard reasoning like AIME 24 and  Dynamic Skipping fails on code generation. ZEDA is the only method preserving competitive performance uniformly across all domains.
Finally, ZEDA achieves average $r_{ZE}$ values of 51.2\% on Qwen and 53.0\% on GLM, exceeding or matching the baselines, indicating that ZEDA attains better performance with comparable or lower computation.

\begin{table*}[h]
\definecolor{mygray}{rgb}{0.85,0.85,0.85}
\definecolor{bluishyellow}{rgb}{0.84, 0.92, 0.85}
\centering
\caption{Performance of ZEDA and baselines on Qwen3-30B-A3B and GLM-4.7-Flash.}
\label{tab:main-results}
\resizebox{\textwidth}{!}{
\begin{tabular}{lcc|ccccc|cccc|cc}
\toprule
\multirow{2}{*}[-3pt]{\textbf{Method}} & \multirow{2}{*}[-3pt]{\shortstack{\textbf{Avg}\\\textbf{Acc}}} & \multirow{2}{*}[-3pt]{\shortstack{\textbf{Avg}\\\textbf{$r_{ZE}$}}} & \multicolumn{5}{c|}{\textbf{Math}} & \multicolumn{4}{c|}{\textbf{Code}} & \multicolumn{2}{c}{\textbf{IF}} \\
\cmidrule(lr){4-8} \cmidrule(lr){9-12} \cmidrule(lr){13-14}
 & & & \textbf{AIME 24} & \textbf{AIME 25} & \textbf{AIME 26} & \textbf{GSM8k} & \textbf{MATH-500}
 & \textbf{LCB v5} & \textbf{LCB v6} & \textbf{HumanEval+} & \textbf{MBPP+}
 & \textbf{IFBench} & \textbf{IFEval} \\
\midrule
\rowcolor{mygray!100}  \textbf{Qwen3-30B-A3B} & $74.9$ & $0.0$ & $80.9$ & $71.0$ & $72.3$ & $95.4$ & $94.4$ & $61.5$ & $57.1$ & $85.6$ & $79.2$ & $39.7$ & $86.3$ \\
  AdaMoE & $54.8$ & $51.9$ & $25.0$ & $24.8$ & $36.7$ & $92.4$ & $79.8$ & $36.1$ & $34.3$ & $80.3$ & $72.8$ & $38.7$ & $82.4$ \\
  Dynamic Skipping & $68.1$ & $43.8$ & $78.1$ & $\underline{67.9}$ & $\textbf{72.5}$ & $95.2$ & $94.4$ & $57.3$ & $51.9$ & $59.1$ & $70.0$ & $32.0$ & $70.4$ \\
  \tableDashLine{14}
  NET$_{\text{SFT}}$ & $72.3$ & $50.0$ & $76.8$ & $65.7$ & $72.1$ & $94.7$ & $94.0$ & $56.5$ & $50.9$ & $86.7$ & $\underline{78.2}$ & $37.7$ & $82.4$ \\
  NET$_{\text{SFT} \rightarrow \text{OPD}}$ & $73.0$ & $50.0$ & $\textbf{79.5}$ & $67.6$ & $70.6$ & $\underline{95.4}$ & $\underline{94.6}$ & $57.0$ & $\underline{52.9}$ & $\underline{87.4}$ & $77.5$ & $38.7$ & $81.7$ \\
  ZEDA$_{\text{SFT}}$ & $\underline{73.3}$ & $51.5$ & $78.1$ & $66.2$ & $71.2$ & $94.8$ & $94.4$ & $\textbf{58.2}$ & $52.8$ & $86.8$ & $\textbf{78.6}$ & $\underline{39.7}$ & $\textbf{85.2}$ \\
\rowcolor{lightblue!100}  ZEDA & $\textbf{74.2}$ & $51.2$ & $\underline{79.0}$ & $\textbf{69.1}$ & $\textbf{72.5}$ & $\textbf{95.5}$ & $\textbf{95.2}$ & $\textbf{58.2}$ & $\textbf{53.2}$ & $\textbf{88.5}$ & $\underline{78.2}$ & $\textbf{42.3}$ & $\underline{84.3}$ \\
\midrule
\rowcolor{mygray!100}  \textbf{GLM-4.7-Flash} & $72.5$ & $0.0$ & $84.2$ & $76.5$ & $74.0$ & $95.2$ & $96.4$ & $48.0$ & $44.4$ & $89.0$ & $75.7$ & $47.3$ & $67.3$ \\
  AdaMoE & $57.1$ & $47.0$ & $44.1$ & $42.4$ & $47.3$ & $93.9$ & $86.4$ & $26.4$ & $28.6$ & $82.7$ & $69.5$ & $43.0$ & $63.8$ \\
  Dynamic Skipping & $67.8$ & $37.5$ & $\textbf{79.9}$ & $69.9$ & $\textbf{74.8}$ & $93.8$ & $\textbf{96.0}$ & $32.3$ & $32.4$ & $86.3$ & $71.6$ & $45.3$ & $63.4$ \\
  \tableDashLine{14}
  NET$_{\text{SFT}}$ & $70.6$ & $50.0$ & $78.2$ & $71.4$ & $68.3$ & $94.1$ & $95.0$ & $\underline{50.6}$ & $44.7$ & $86.8$ & $\underline{74.2}$ & $\underline{47.0}$ & $65.8$ \\
  NET$_{\text{SFT} \rightarrow \text{OPD}}$ & $\underline{70.9}$ & $50.0$ & $78.6$ & $\underline{71.8}$ & $72.9$ & $94.2$ & $\underline{95.6}$ & $49.5$ & $43.8$ & $\underline{87.1}$ & $\textbf{74.3}$ & $46.7$ & $65.1$ \\
  ZEDA$_{\text{SFT}}$ & $\underline{70.9}$ & $52.8$ & $78.1$ & $71.4$ & $71.9$ & $\textbf{95.2}$ & $95.0$ & $49.9$ & $\underline{45.0}$ & $\textbf{88.1}$ & $72.3$ & $46.7$ & $\underline{66.4}$ \\
\rowcolor{lightblue!100}  ZEDA & $\textbf{71.8}$ & $53.0$ & $\underline{79.8}$ & $\textbf{73.1}$ & $\underline{74.4}$ & $\underline{94.4}$ & $95.2$ & $\textbf{51.6}$ & $\textbf{45.6}$ & $86.3$ & $73.5$ & $\textbf{47.3}$ & $\textbf{68.2}$ \\
\bottomrule
\end{tabular}
}
\end{table*}

\paragraph{Adaptation Time.}
Table~\ref{tab:training time} reports the training time of the ZEDA pipeline. 
ZEDA requires less than 31 hours for Qwen3-30B-A3B and 62 hours for GLM-4.7-Flash on 8 H200 GPUs, which is negligible compared with prior MoE pre-training and post-training costs, demonstrating its cost-effectiveness.

\begin{table}[h]
\centering
\caption{Adaptation time (hours) of Qwen3-30B-A3B and GLM-4.7-Flash on 8 NVIDIA H200 GPUs}
\label{tab:training time}
\resizebox{\textwidth}{!}{
\begin{tabular}{lcccc|lcccc}
\toprule
\multirow{2}{*}[-3pt]{\textbf{Qwen3-30B-A3B}} & \textbf{All} & \textbf{SFT Data Rollout} & \textbf{SFT} & \textbf{OPD} & \multirow{2}{*}[-3pt]{\textbf{GLM-4.7-Flash}} & \textbf{All} & \textbf{SFT Data Rollout} & \textbf{SFT} & \textbf{OPD} \\
\cmidrule(lr){2-5} \cmidrule(lr){7-10}
 & 30.12 & 8.16 & 1.97 & 19.99 &  & 61.37 & 14.56 & 4.51 & 42.30 \\
\bottomrule
\end{tabular}%
}
\end{table}

\subsection{Inference Efficiency}

ZEDA yields average zero-expert activation ratios ($r_{\text{ZE}}$) of 51.2\% and 53.0\% on Qwen and GLM respectively, effectively halving expert-level computation. We further demonstrate the practical inference speedups achieved by the resulting dynamic MoE.

Inference efficiency is evaluated by comparing the original model with its ZEDA-adapted counterpart at $8192$ sequence length, using SGLang~\citep{zheng2024sglang} as the inference framework with the maximum concurrency set to 32. 
We randomly sample 256 examples from the training data to construct the test set. To ensure a fair comparison across models, for each target sequence length we control the total numbers of input and output tokens to be identical across compared models and to match the intended test sequence length. In addition, the input sequence content is kept exactly the same across models. We report the throughput results on 1 $\times$ H200 GPU.
We measure both prefill and decode efficiency, and we also provide the theoretical analysis of inference efficiency in Appendix~\ref{app:flop}.

As shown in Figure~\ref{tab:inference-efficiency}, ZEDA delivers consistent inference gains across both backbone models, achieving approximately 20\% speedup during the prefill and decode phases, demonstrating its effectiveness in improving model's inference efficiency.


\begin{figure}[h]
  \centering
  \includegraphics[width=\textwidth]{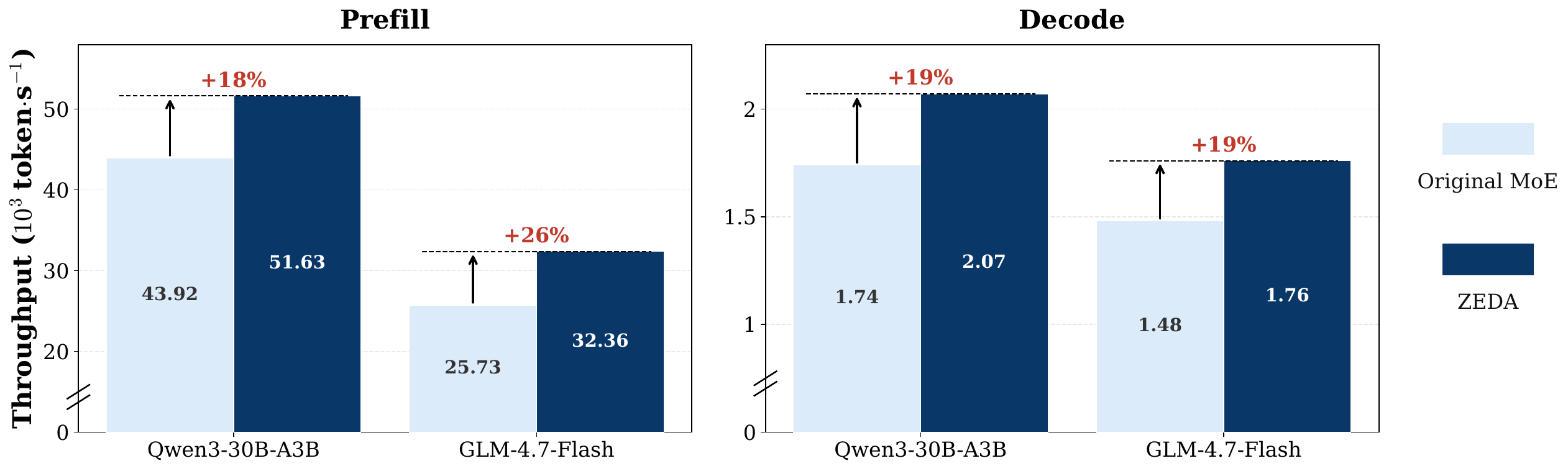}
  \caption{Inference efficiency comparison between the original MoE and ZEDA at $8192$ sequence length. Speedup is defined relative to the original MoE.}
  \label{tab:inference-efficiency}
\end{figure}

\section{Analysis}
We provide a detailed analysis of the dynamic characteristics of zero expert activation ($\S$~\ref{sec:analysis_rze}), the effects of different adaptation durations ($\S$~\ref{sec:adaptation-cost}), ablation studies on zero-expert group weight $w$($\S$~\ref{sec:analyze_effect_of_weight}), $\mathcal{L}_{GA}$ coefficient $\alpha$($\S$~\ref{sec:ablation}), training stages($\S$~\ref{sec:training-stages}), and router probability renormalization($\S$~\ref{sec:renorm}), and ZEDA’s performance on OOD tasks ($\S$~\ref{sec:ood}).

\subsection{Zero Expert Activation Dynamics}
\label{sec:analysis_rze}

ZEDA transforms a static MoE model into a dynamic one in which different tokens exhibit different $r_{ZE}$ values, corresponding to varying computation amounts. This section provides a deeper investigation into this token-level dynamism, using Qwen3-30B-A3B. The analysis examines how $r_{ZE}$ relates to distillation signals, response patterns, task difficulty, and layer-wise behavior, aiming to establish connections between computation allocation in the dynamic MoE and other interpretable metrics.

\begin{figure}[h]
  \centering
  \includegraphics[width=\textwidth]{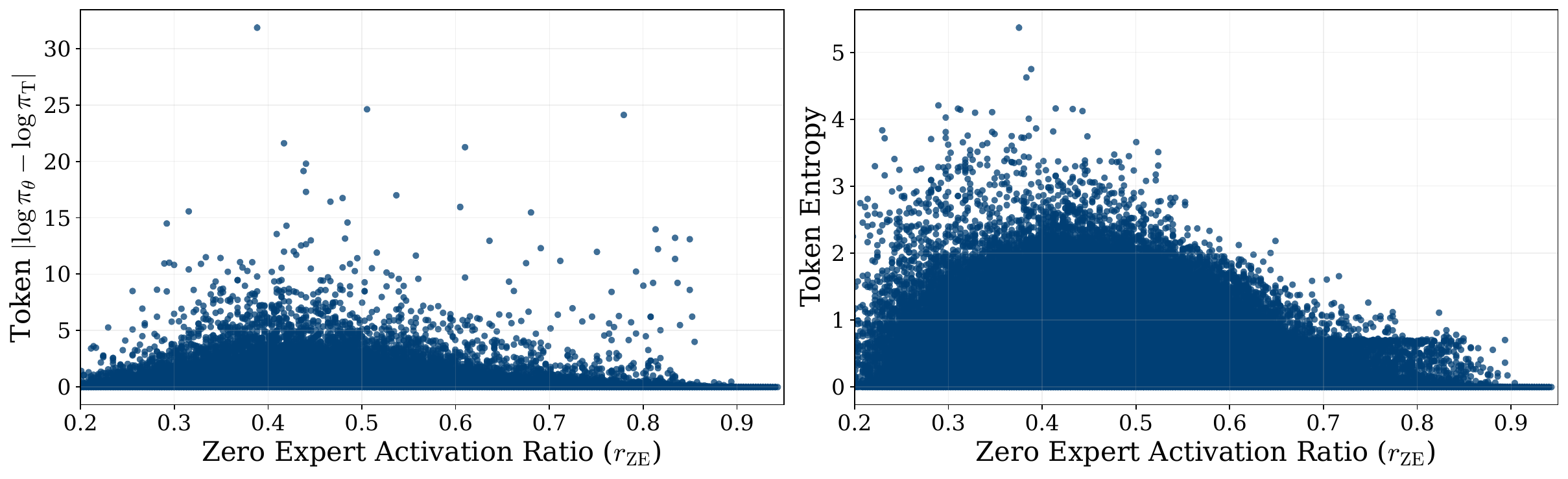}
  \caption{Distribution of token-level $r_{ZE}$ versus teacher-student logp-diff (left) and entropy (right) for all generated tokens across 110 rollouts. Each point represents a single token.}
  \label{fig:entropy}
\end{figure}

\paragraph{Teacher-Student Logp-Diff and Entropy.}
To analyze factors affecting token-level $r_{ZE}$, 110 prompts (10 per benchmark) are sampled and decoded with the ZEDA-adapted dynamic MoE model. For each generated token, we record the student log probability $\log \pi_\theta(y_t \mid x, y_{<t})$ and entropy, and compute the teacher log probability $\log \pi_T(y_t \mid x, y_{<t})$ on the same token to obtain the teacher-student logp-diff $\Delta_{\text{logp}}$. Figure~\ref{fig:entropy} visualizes all tokens from the 110 prompts. Tokens with larger $\Delta_{\text{logp}}$ or higher entropy tend to have lower $r_{ZE}$, clustering in the upper-left. the dynamic MoE intrinsically allocates more computation, i.e., activates fewer zero experts, when the teacher-student distributional gap or model uncertainty is larger.

\begin{figure}[h]
  \centering
  \includegraphics[width=\textwidth]{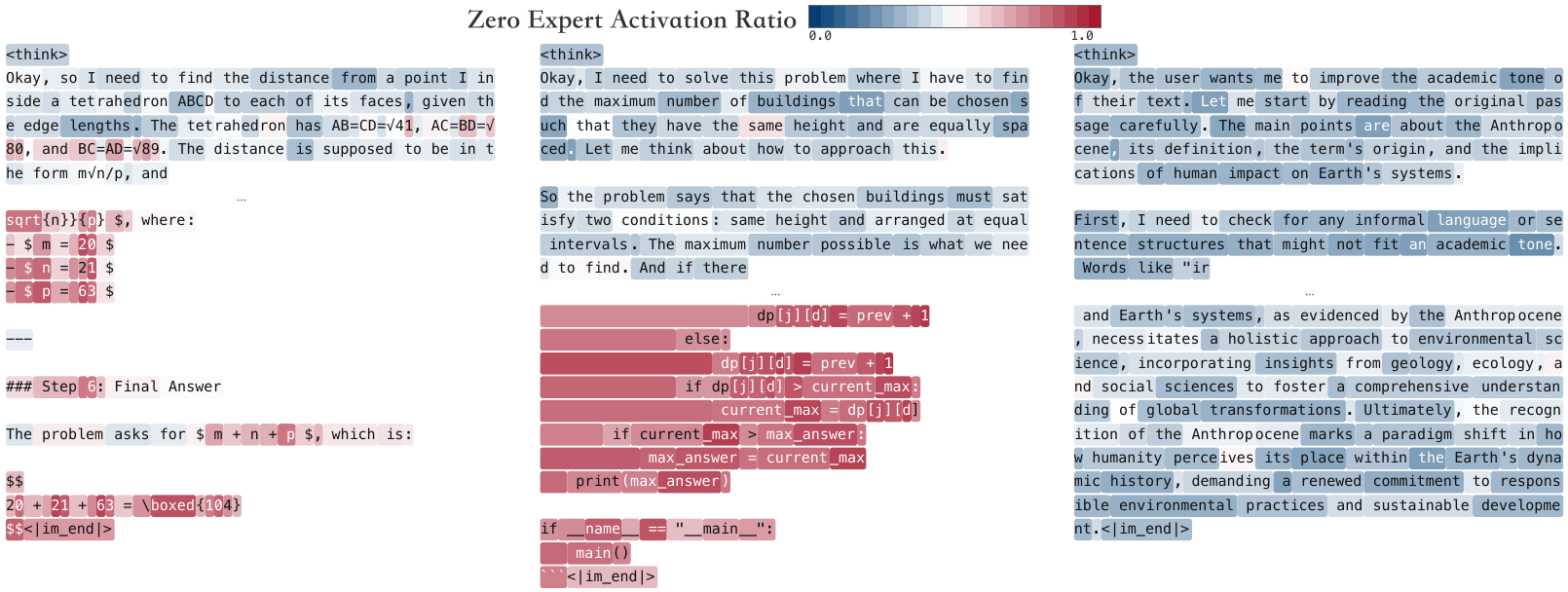}
  \caption{Visualization of per-token $r_{ZE}$ for decoded text, showing one sampled response from AIME24 (left), LiveCodeBench v5 (middle), and IFBench (right), respectively. Due to space constraints, only the first and last 80 tokens of each response are retained.}
  \label{fig:pattern}
\end{figure}

\begin{figure}[h]
  \centering
  \includegraphics[width=\textwidth]{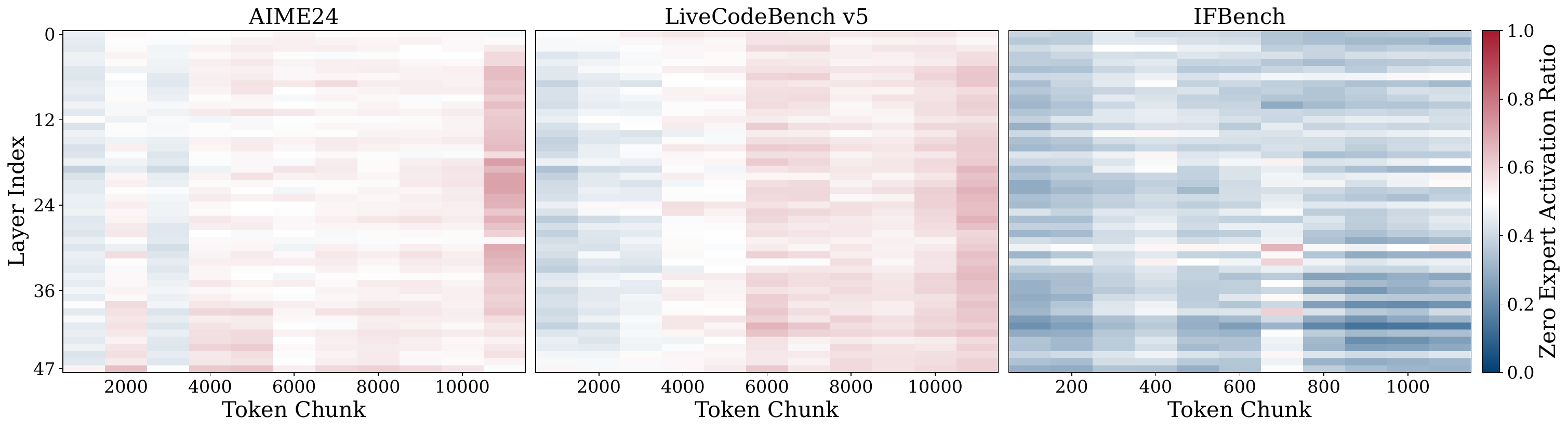}
  \caption{Visualization of $r_{ZE}$ across layers and response positions, using the same data in Figure~\ref{fig:pattern}. Token-level $r_{ZE}$ values are averaged over chunks of size 1000 (for AIME24 and LiveCodeBench v5) and 100 (for IFBench), respectively. The last chunk is averaged over its actual token number.}
  \label{fig:layer}
\end{figure}

\paragraph{Response Pattern.}
Aligning per-token $r_{ZE}$ of the 110 sampled responses with the decoded text reveals a clear relationship between $r_{ZE}$ and response pattern. Figure~\ref{fig:pattern} presents 3 representative examples. Compared with natural text, code fragments and mathematical expression exhibit notably higher $r_{ZE}$, indicating that the model intrinsically assigns less computation to these structured segments. 
Since math and code rollouts often contain many such segments after the thinking process, their average $r_{ZE}$ tends to increase toward the response end, while instruction-following responses show a more uniform $r_{ZE}$ distribution, as illustrated in Figure~\ref{fig:layer}.

\paragraph{Task Difficulty.}
The relationship between $r_{ZE}$ and task difficulty is further investigated. Table~\ref{tab:difficulty} reports the $r_{ZE}$ and performance of ZEDA on MATH-500, which provides human-annotated difficulty levels, and on AIME24, a generally considered more challenging task. ZEDA achieves comparable performance and $r_{ZE}$ across all five difficulty levels of MATH-500, and the corresponding $r_{ZE}$ values remain close to those observed on AIME24. This suggests that $r_{ZE}$ is largely independent of task difficulty. The model adjusts computation allocation based on the token-level characteristics within a single response rather than the overall difficulty of the task itself.

\begin{table*}[h]
\centering
\caption{Performance and $r_{ZE}$ of ZEDA on five difficulty level tasks of MATH-500 and AIME24.}
\label{tab:difficulty}
\resizebox{0.85\textwidth}{!}{
\begin{tabular}{l|ccccc|c}
\toprule
\multirow{2}{*}{\textbf{}} & \multicolumn{5}{c|}{\textbf{MATH-500}} & \multirow{2}{*}{\textbf{AIME 24}} \\
\cmidrule(lr){2-6}
 & \textbf{Level 1} & \textbf{Level 2} & \textbf{Level 3} & \textbf{Level 4} & \textbf{Level 5} & \\
\midrule
  Performance ($r_{ZE}$) & $95.3$ ($51.1$) & $95.6$ ($51.8$) & $97.1$ ($52.2$) & $94.5$ ($52.5$) & $94.0$ ($52.5$) & $79.0$ ($52.1$) \\
\bottomrule
\end{tabular}
}
\end{table*}

\paragraph{Layer.}
For each of the 110 responses, $r_{ZE}$ on the 48 MoE layers of the dynamic model is computed. 
Figure~\ref{fig:layer} presents the layer-wise $r_{ZE}$ distributions for 3 representative cases. Although minor variations exist across layers, the differences are relatively small and exhibit no systematic pattern.

\paragraph{Connecting the Observations.}
The above analyses reveal that $r_{ZE}$ is uncorrelated with task difficulty yet strongly related to teacher-student logp-diff. This can be explained by the nature of the self-distillation training data. Diverse sources of the training data make sample-level accuracy-based difficulty signals generally unavailable. In contrast, larger $\Delta_{\text{logp}}$ directly implies larger $\mathcal{L}_{\mathrm{SFT}}$ (Eq.~\ref{eq:sft}) and $\mathcal{L}_{\mathrm{OPD}}$ (Eq.~\ref{eq:opd_loss}). When these task losses dominate, the relative influence of $\mathcal{L}_{\mathrm{GA}}$, which encourages higher $r_{ZE}$, diminishes, leading to lower $r_{ZE}$ for such tokens. Furthermore, the correlations of $r_{ZE}$ with entropy and response patterns align with prior findings. Tokens with higher student-model entropy tend to exhibit larger $\Delta_{\text{logp}}$~\citep{ko2026scaling}, and low-entropy tokens are often code or math expressions~\citep{wang2025beyond}.

\subsection{Effect of Adaptation Cost}
\label{sec:adaptation-cost}

\begin{figure}[h]
  \centering
  \includegraphics[width=\textwidth]{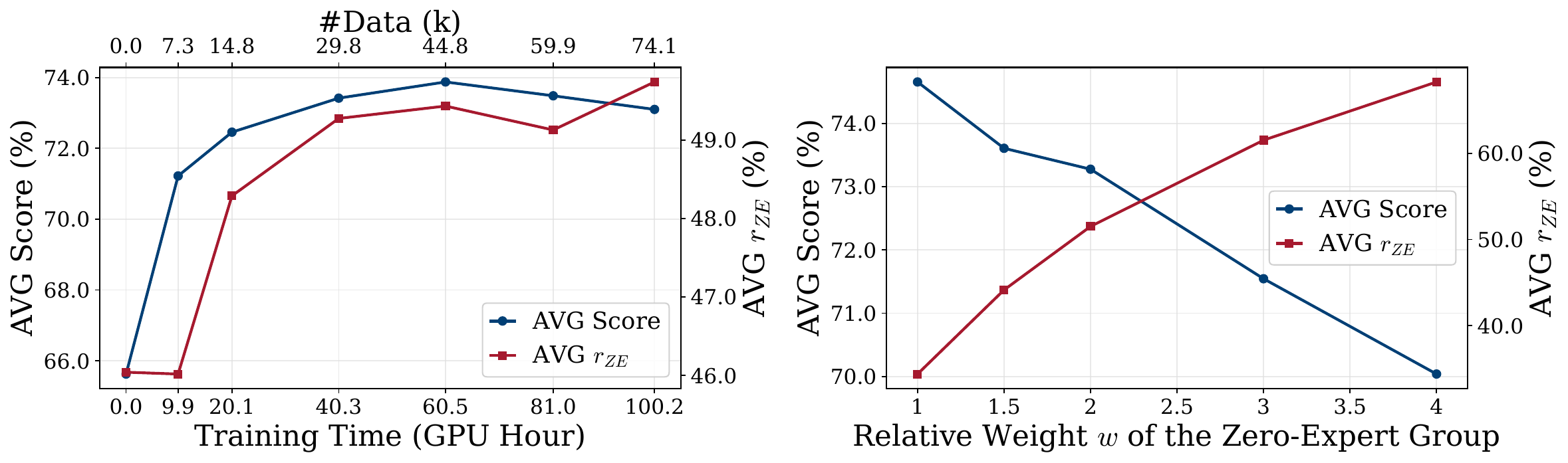}
  \caption{Scaling trend of average benchmark score and $r_{ZE}$ over different amounts of SFT data and adaptation time (left). Effect of the zero-expert group weight on average score and $r_{ZE}$ (right). }
  \label{fig:effect-transfer-zcerate-combined}
\end{figure}

To study the scaling trend of zero expert adaptation, we track the average benchmark score and $r_{ZE}$ throughout the SFT stage.
As illustrated in Figure~\ref{fig:effect-transfer-zcerate-combined} (left), both metrics exhibit similar evolution as the amount of SFT data and GPU hours increases, which is a rapid initial ascent followed by convergence at approximately 60k prompts. This trend indicates that the majority of useful adaptation happens relatively early, when the router is learning how to incorporate the injected zero experts while preserving the backbone's original capabilities. After this point, additional supervised adaptation mainly provides incremental refinements.

These empirical results justify our 60k self-distillation prompt setting, as this scale achieves a performance plateau and stable routing patterns. The observed "dual saturation" underscores the sample efficiency of ZEDA, demonstrating that the modified architecture can reach a stable, high-performance state with affordable post-training costs.

\subsection{Ablation Studies on ZEDA Design}

\subsubsection{Effect of $w$ and $r_{ZE}$}
\label{sec:analyze_effect_of_weight}

To investigate the impact of group-level balancing strength and the zero expert activation ratio, we vary the zero-expert group weight $w$ and analyze its effect on model performance and routing.
As shown in Figure~\ref{fig:effect-transfer-zcerate-combined} (right), increasing $w$ monotonically elevates $r_{ZE}$ but leads to a gradual decline in benchmark scores. This confirms that $w$ serves as an effective control knob for the quality-efficiency trade-off in ZEDA.
Empirical results indicate that $w{=}2$ offers the optimal balance, yielding significantly higher zero-expert utilization than $w{\in}\{1, 1.5\}$ while maintaining competitive performance. In contrast, pushing $w$ further to 3 or 4 causes a more pronounced accuracy drop. Consequently, we select $w{=}2$ as the preferred operating point.

\subsubsection{$\mathcal{L}_{GA}$ Coefficient $\alpha$ Ablation}
\label{sec:ablation}

\begin{table*}[h]
\centering
\caption{ZEDA$_{\text{SFT}}$ performance on Qwen3-30B-A3B with different $\alpha$.}
\label{tab:alpha}
\resizebox{\textwidth}{!}{
\begin{tabular}{lcc|ccccc|cccc|cc}
\toprule
\multirow{2}{*}{$\mathbf{\alpha}$} & \textbf{Avg} & \textbf{Avg} & \multicolumn{5}{c|}{\textbf{Math}} & \multicolumn{4}{c|}{\textbf{Code}} & \multicolumn{2}{c}{\textbf{IF}} \\
\cmidrule(lr){4-8} \cmidrule(lr){9-12} \cmidrule(lr){13-14}
 &\textbf{Acc} &\textbf{$r_{CE}$} & \textbf{AIME 24} & \textbf{AIME 25} & \textbf{AIME 26} & \textbf{GSM8k} & \textbf{MATH-500}
 & \textbf{LCB v5} & \textbf{LCB v6} & \textbf{HumanEval+} & \textbf{MBPP+}
 & \textbf{IFBench} & \textbf{IFEval} \\
\midrule
  $0.001$ & $74.3$ & $31.8$ & $79.2$ & $68.5$ & $73.1$ & $95.2$ & $94.8$ & $59.7$ & $55.6$ & $88.3$ & $79.0$ & $38.0$ & $86.3$ \\
  $0.01$ & $73.6$ & $44.6$ & $78.3$ & $67.8$ & $71.7$ & $95.2$ & $94.4$ & $59.0$ & $52.5$ & $87.1$ & $78.4$ & $41.7$ & $84.1$ \\
  $0.1$ & $73.3$ & $51.5$ & $78.1$ & $66.2$ & $71.3$ & $94.8$ & $94.4$ & $58.2$ & $52.8$ & $86.8$ & $78.6$ & $39.7$ & $85.2$ \\
  $1.0$ & $73.3$ & $50.9$ & $77.4$ & $67.3$ & $69.8$ & $94.9$ & $94.8$ & $59.0$ & $54.4$ & $87.0$ & $78.7$ & $39.7$ & $83.6$ \\
\bottomrule
\end{tabular}
}
\end{table*}

The loss coefficient $\alpha$ in Eq.~\ref{eq:gal} controls how strongly the Group Auxiliary Loss $\mathcal{L}_{GA}$ influences the overall training objective. To investigate its effect, the SFT stage of ZEDA is conducted on Qwen3-30B-A3B with $\alpha$ varied across \{0.001, 0.01, 0.1, 1.0\}, while the relative weight $w$ is fixed at 2, corresponding to a target $r_{ZE}$ of 50\%.
As shown in Table~\ref{tab:alpha},
at $\alpha{=}0.1$, the observed $r_{ZE}$ is closest to the 50\% target prescribed by $\mathcal{L}_{GA}$, while the average accuracy remains comparable to the original model. This suggests that $\alpha{=}0.1$ achieves the best effect of enforcing the intended zero expert utilization. Based on these findings, $\alpha$ is set to 0.1 in all subsequent experiments.

\subsubsection{Effect of Training Stages}
\label{sec:training-stages}

The full ZEDA pipeline employs a two-stage self-distillation strategy. To assess whether both stages are necessary, three variants are compared: (1) SFT only, (2) OPD only, and (3) the full SFT$\to$OPD pipeline. To ensure a fair comparison, the SFT-only and OPD-only variants are trained with an increased number of steps so that their total computational cost matches or exceeds that of the full pipeline.
As shown in Table~\ref{tab:training-stage}, the full SFT$\to$OPD pipeline consistently outperforms both single-stage alternatives, with OPD alone performing the worst. 
This may be because SFT first establishes stable zero-expert routing patterns, without which OPD must simultaneously learn routing decisions and generate coherent responses, compounding the adaptation difficulty. Once SFT has stabilized the router, OPD can focus on closing the remaining distribution gap under on-policy rollouts, yielding further gains that neither stage achieves in isolation.

\begin{table*}[h]
\centering
\caption{Performance of ZEDA with different self-distillation strategies on Qwen3-30B-A3B.}
\label{tab:training-stage}
\resizebox{\textwidth}{!}{
\begin{tabular}{lcc|ccccc|cccc|cc}
\toprule
\multirow{2}{*}{} &\multirow{2}{*}[-3pt]{\shortstack{\textbf{Avg}\\\textbf{Acc}}} & \multirow{2}{*}[-3pt]{\shortstack{\textbf{Avg}\\\textbf{$r_{ZE}$}}} & \multicolumn{5}{c|}{\textbf{Math}} & \multicolumn{4}{c|}{\textbf{Code}} & \multicolumn{2}{c}{\textbf{IF}} \\
\cmidrule(lr){4-8} \cmidrule(lr){9-12} \cmidrule(lr){13-14}
 & & & \textbf{AIME 24} & \textbf{AIME 25} & \textbf{AIME 26} & \textbf{GSM8k} & \textbf{MATH-500}
 & \textbf{LCB v5} & \textbf{LCB v6} & \textbf{HumanEval+} & \textbf{MBPP+}
 & \textbf{IFBench} & \textbf{IFEval} \\
\midrule
  SFT & $73.6$ & $51.5$ & $78.5$ & $\mathbf{69.3}$ & $69.8$ & $95.3$ & $94.8$ & $58.1$ & $52.1$ & $\mathbf{88.6}$ & $79.0$ & $40.3$ & $83.5$ \\
  OPD & $72.9$ & $50.1$ & $78.4$ & $67.2$ & $72.3$ & $94.9$ & $94.4$ & $57.0$ & $51.5$ & $87.3$ & $\mathbf{79.6}$ & $36.3$ & $83.2$ \\
  SFT $\rightarrow$ OPD & $\mathbf{74.2}$ & $51.2$ & $\mathbf{79.0}$ & $69.1$ & $\mathbf{72.5}$ & $\mathbf{95.5}$ & $\mathbf{95.2}$ & $\mathbf{58.2}$ & $\mathbf{53.2}$ & $88.5$ & $78.2$ & $\mathbf{42.3}$ & $\mathbf{84.3}$ \\
\bottomrule
\end{tabular}
}
\end{table*}

\subsubsection{Impact of Router Probability Renormalization}
\label{sec:renorm}

\begin{table*}[b]
\centering
\caption{Performance of ZEDA$_{\text{SFT}}$ with and without renormalization on Qwen3-30B-A3B.}
\label{tab:renorm}
\resizebox{\textwidth}{!}{
\begin{tabular}{lcc|ccccc|cccc|cc}
\toprule
\multirow{2}{*}{} & \multirow{2}{*}[-3pt]{\shortstack{\textbf{Avg}\\\textbf{Acc}}} & \multirow{2}{*}[-3pt]{\shortstack{\textbf{Avg}\\\textbf{$r_{ZE}$}}} & \multicolumn{5}{c|}{\textbf{Math}} & \multicolumn{4}{c|}{\textbf{Code}} & \multicolumn{2}{c}{\textbf{IF}} \\
\cmidrule(lr){4-8} \cmidrule(lr){9-12} \cmidrule(lr){13-14}
 & & & \textbf{AIME 24} & \textbf{AIME 25} & \textbf{AIME 26} & \textbf{GSM8k} & \textbf{MATH-500}
 & \textbf{LCB v5} & \textbf{LCB v6} & \textbf{HumanEval+} & \textbf{MBPP+}
 & \textbf{IFBench} & \textbf{IFEval} \\
\midrule
  w/ Renorm  & $71.6$ & $51.0$ & $76.6$ & $64.8$ & $67.9$ & $94.8$ & $94.2$ & $55.0$ & $48.6$ & $\mathbf{87.4}$ & $77.7$ & $38.3$ & $82.1$ \\
  w/o Renorm & $\mathbf{73.3}$ & $51.5$ & $\mathbf{78.1}$ & $\mathbf{66.2}$ & $\mathbf{71.3}$ & $94.8$ & $\mathbf{94.4}$ & $\mathbf{58.2}$ & $\mathbf{52.8}$ & $86.8$ & $\mathbf{78.6}$ & $\mathbf{39.7}$ & $\mathbf{85.2}$\\
\bottomrule
\end{tabular}
}
\end{table*}

As defined in Eq.~\ref{eq:dynamic_moe}, the dynamic MoE output after zero-expert injection is $\tilde{y}(h) = \sum_{i \in \tilde{\mathcal{S}}(h) \cap \mathcal{E}} \tilde{g}_i(h)\, E_i(h)$. In this formulation, the routing weights $\tilde{g}_i(h)$ of the remaining normal experts are not renormalized after zero experts are removed from the top-$K$ selection. An alternative is to redistribute the router probability among the active normal experts through renormalization. Concretely, the renormalized output becomes
\begin{equation}
\label{eq:renorm}
\tilde{y}_{\text{renorm}}(h) = \sum_{i \in \tilde{\mathcal{S}}(h) \cap \mathcal{E}} \frac{\tilde{g}_i(h)}{\sum_{j \in \tilde{\mathcal{S}}(h) \cap \mathcal{E}} \tilde{g}_j(h)} \, E_i(h),
\end{equation}
where the routing weights are rescaled to sum to one over the active normal experts.

To evaluate this design choice, SFT is conducted on Qwen3-30B-A3B with and without renormalization under identical hyperparameters. As reported in Table~\ref{tab:renorm}, renormalization leads to a consistent accuracy drop compared with the default formulation. A likely reason is that in the original model, the sum of routing weights over the top-$K$ experts is calibrated during pre-training to produce outputs at a certain magnitude. And renormalization artificially amplifies the routing weights of the active normal experts, inflating the effective scale of the MoE residual branch.

\subsection{Out-of-Distribution Generalization}
\label{sec:ood}

To evaluate whether zero-expert adaptation preserves capability beyond the in-distribution evaluation suite, we further test all methods on two out-of-distribution (OOD) benchmarks, MMLU-Redux~\citep{gema2025we} and GPQA-Diamond~\citep{rein2023gpqa}. These benchmarks primarily assess knowledge-intensive question answering and scientific reasoning, which are out of distribution with respect to the math, code, and instruction-following domains represented in the self-distillation training data. Unless otherwise specified, the evaluation setup follows Section~\ref{sec:settings}. Deviations are limited to a maximum generation length of 32k tokens, avg@8 for GPQA-Diamond, and avg@1 for MMLU-Redux.
Table~\ref{tab:ood-results} shows that ZEDA consistently preserves competitive OOD accuracy while maintaining high zero-expert utilization (47.2\% and 50.0\% average $r_{ZE}$, respectively) on both Qwen3-30B-A3B and GLM-4.7-Flash, indicating a favorable quality-efficiency trade-off under distribution shift. These results further demonstrate the strong out-of-distribution generalization capability of ZEDA.

\begin{table}[h]
\definecolor{mygray}{rgb}{0.85,0.85,0.85}
  \centering
  \caption{OOD generalization results for Qwen3-30B-A3B (left) and GLM-4.7-Flash (right) on MMLU-Redux (MMLU) and GPQA-Diamond (GPQA).}
  \label{tab:ood-results}
  \scriptsize
  \resizebox{\textwidth}{!}{
  \begin{tabular}{@{}lcc|cc|lcc|cc@{}}
  \toprule
  \textbf{Method} 
  & \textbf{Avg Acc} & \textbf{Avg $r_{ZE}$} 
  & \textbf{MMLU} & \textbf{GPQA}
  & \textbf{Method} 
  & \textbf{Avg Acc} & \textbf{Avg $r_{ZE}$} 
  & \textbf{MMLU} & \textbf{GPQA} \\
  \midrule
  \rowcolor{mygray!100}
  \textbf{Qwen3-30B-A3B} & $76.7$ & $0.0$ & $90.1$ & $63.3$
  & \textbf{GLM-4.7-Flash} & $76.1$ & $0.0$ & $89.8$ & $62.4$ \\

  ZEDA & $76.2$ & $47.2$ & $89.2$ & $63.2$
  & ZEDA & $72.9$ & $50.0$ & $89.0$ & $56.8$ \\
  \bottomrule
  \end{tabular}}
\end{table}

\section{Related Work}

\subsection{Dynamic Expert Activation in Mixture-of-Experts LLMs}
Mixture-of-Experts (MoE) has emerged as an effective architecture for scaling large language models by increasing model capacity while keeping bounded per-token computation~\citep{shazeer2017outrageously, lepikhin2020gshard, fedus2022switch, du2022glam}. Subsequent studies~\citep{zoph2022st, dai2024deepseekmoe, jiang2024mixtral} further improved sparse expert training and specialization, making MoE a practical design for large-scale language modeling. Nevertheless, in standard MoE architectures, routing is typically constrained by a fixed top-$k$ policy, meaning that while expert activation is input-dependent, the computation budget remains largely static across tokens.

To address this limitation, prior work has mainly proceeded along two directions. One line of work improves efficiency by reducing expert redundancy or activated computation, and can be further categorized into experts pruning~\citep{lu2024not, liu2024efficient}, merging~\citep{li2023merge, chen2024retraining} and compression~\citep{li2023merge,chen2025eac,zhang2025diversifying,hao2026lightmoe}. The other direction replaces the static top-$k$ routing policy with dynamic expert activation, enabling token-level input-dependent allocation of computation budgets~\citep{lu2024not, jin2024moe++, zhou2022mixture, huang2024harder, zeng2024adamoe, yue2024ada, guo2024dynamic, sun2026expert}. Early work relaxed the fixed-cardinality assumption by allowing the number of activated experts to vary across tokens, either through expert-selected token assignment~\citep{zhou2022mixture} or by allocating more experts to harder inputs~\citep{huang2024harder}. More recent work adapts these methods to modern autoregressive MoE language models: AdaMoE~\citep{zeng2024adamoe} introduces null experts so that the number of real activated experts can vary with minimal changes to standard routing, Ada-K Routing~\citep{yue2024ada} explicitly learns a token-dependent $k$ for expert routing, DynMoE~\citep{guo2024dynamic} jointly auto-tunes both the total number of experts and the per-token activation budget, and Expert Threshold Routing~\citep{sun2026expert} replaces fixed top-$k$ selection with threshold-based activation to obtain causal variable-size expert sets with improved load balancing, MoE++~\citep{jin2024moe++} extends dynamic routing into dynamic computation-path selection by introducing zero-computation experts, which allow some tokens to bypass expensive FFN computation. Dynamic activation can also be introduced from a deployment perspective via inference-time expert skipping~\citep{lu2024not}, where selected experts are conditionally bypassed at inference time without fundamentally changing the underlying router.

In contrast to these approaches, we study a lower-cost form of dynamic expert activation that begins at the post-training stage, instead of relying on expensive re-pretraining or substantial router redesign. Our method operates entirely in the post-training regime and follows the zero-computation expert paradigm of MoE++~\citep{jin2024moe++}, which has also been validated at industrial scale in Meituan's LongCat-Flash~\citep{team2025longcat}. This design avoids substantive architectural modifications to the underlying MoE model, making it particularly appealing for practical adaptation and deployment.

\subsection{Self-Distillation}

Knowledge distillation (KD) was originally introduced as a teacher–student framework in which a student network learns from the softened output distribution of a stronger teacher network~\citep{hinton2015distilling}. This paradigm had already been broadly extended to language modeling, from sequence-level distillation in neural text generation \citep{kim2016sequence}, supervised distillation in autoregressive language models~\citep{sanh2019distilbert} to more recent rationale-augmented distillation with large language models \citep{hsieh2023distilling}. More recently, on-policy distillation methods for language models argued that standard distillation is inherently off-policy, since the student is trained on teacher-generated trajectories but tested on its own generations. To reduce this mismatch, methods such as MiniLLM~\citep{gu2023minillm} and GKD~\citep{agarwal2024policy} apply teacher supervision on student-sampled sequences, a perspective that has also been highlighted in recent practitioner discussions of language model post-training~\citep{lu2025onpolicydistillation}. In parallel, self-distillation has been shown to improve performance even without an external teacher~\citep{furlanello2018born,zhang2019your}. More recent work has further explored on-policy self-distillation and demonstrated its potential in scenarios such as reasoning and continual learning~\citep{zhao2026self,shenfeld2026self,hubotter2026reinforcement}.

Beyond the capability improvement and task-specific adaptation, recent work has also examined self-distillation in the context of architecture adaptation towards higher computational efficiency. RAD~\citep{hoshino2025rad} and HALO~\citep{chen2026hybrid} utilize self-distillation as a principled mechanism to transform standard full-attention layers into computationally efficient alternatives, thereby achieving substantial gains in inference efficiency while maintaining model performance. LaDiMo~\citep{kim2024ladimo} employs layer-wise distillation to transform dense models into sparse MoE architectures, facilitating efficient sparse architecture adaptation. Nevertheless, existing efforts have primarily focused on static architecture conversion, with limited attention to using self-distillation for efficient dynamic MoE architectures. In particular, the introduction of on-policy self-distillation to reduce redundant expert activation in MoE models remains underexplored, representing a critical gap in the current literature.

\section{Conclusion}

This study presents ZEDA, a lightweight and effective framework for migrating post-trained static MoE models to dynamic ones through zero-expert injection and two-stage self-distillation. With the group auxiliary loss, ZEDA regulates computation allocation while preserving the delicate routing distributions of the original MoE. Empirical evaluations across multiple architectures and benchmarks demonstrate that ZEDA eliminates over half the expert computation and provides significant inference speedups with negligible impact on model performance. 
These findings validate that post-trained MoE models can be adapted to efficient dynamic ones via self-distillation, offering a practical solution for enhancing the deployment efficiency of large-scale MoE systems across diverse domains.

\bibliography{Public/paper}

@article{ouyang2022training,
  title={Training language models to follow instructions with human feedback},
  author={Ouyang, Long and Wu, Jeffrey and Jiang, Xu and Almeida, Diogo and Wainwright, Carroll and Mishkin, Pamela and Zhang, Chong and Agarwal, Sandhini and Slama, Katarina and Ray, Alex and others},
  journal={Advances in neural information processing systems},
  volume={35},
  pages={27730--27744},
  year={2022}
}

@article{li2024numinamath,
  title={Numinamath: The largest public dataset in ai4maths with 860k pairs of competition math problems and solutions},
  author={Li, Jia and Beeching, Edward and Tunstall, Lewis and Lipkin, Ben and Soletskyi, Roman and Huang, Shengyi and Rasul, Kashif and Yu, Longhui and Jiang, Albert Q and Shen, Ziju and others},
  journal={Hugging Face repository},
  volume={13},
  number={9},
  pages={9},
  year={2024}
}

@article{jain2024livecodebench,
  title={Livecodebench: Holistic and contamination free evaluation of large language models for code},
  author={Jain, Naman and Han, King and Gu, Alex and Li, Wen-Ding and Yan, Fanjia and Zhang, Tianjun and Wang, Sida and Solar-Lezama, Armando and Sen, Koushik and Stoica, Ion},
  journal={arXiv preprint arXiv:2403.07974},
  year={2024}
}

@article{wang2025beyond,
  title={Beyond the 80/20 rule: High-entropy minority tokens drive effective reinforcement learning for llm reasoning},
  author={Wang, Shenzhi and Yu, Le and Gao, Chang and Zheng, Chujie and Liu, Shixuan and Lu, Rui and Dang, Kai and Chen, Xionghui and Yang, Jianxin and Zhang, Zhenru and others},
  journal={arXiv preprint arXiv:2506.01939},
  year={2025}
}

@article{shazeer2017outrageously,
  title={Outrageously large neural networks: The sparsely-gated mixture-of-experts layer},
  author={Shazeer, Noam and Mirhoseini, Azalia and Maziarz, Krzysztof and Davis, Andy and Le, Quoc and Hinton, Geoffrey and Dean, Jeff},
  journal={arXiv preprint arXiv:1701.06538},
  year={2017}
}

@article{fedus2022switch,
  title={Switch transformers: Scaling to trillion parameter models with simple and efficient sparsity},
  author={Fedus, William and Zoph, Barret and Shazeer, Noam},
  journal={Journal of Machine Learning Research},
  volume={23},
  number={120},
  pages={1--39},
  year={2022}
}

@inproceedings{du2022glam,
  title={Glam: Efficient scaling of language models with mixture-of-experts},
  author={Du, Nan and Huang, Yanping and Dai, Andrew M and Tong, Simon and Lepikhin, Dmitry and Xu, Yuanzhong and Krikun, Maxim and Zhou, Yanqi and Yu, Adams Wei and Firat, Orhan and others},
  booktitle={International conference on machine learning},
  pages={5547--5569},
  year={2022},
  organization={PMLR}
}

@article{zoph2022st,
  title={St-moe: Designing stable and transferable sparse expert models},
  author={Zoph, Barret and Bello, Irwan and Kumar, Sameer and Du, Nan and Huang, Yanping and Dean, Jeff and Shazeer, Noam and Fedus, William},
  journal={arXiv preprint arXiv:2202.08906},
  year={2022}
}

@inproceedings{dai2024deepseekmoe,
  title={Deepseekmoe: Towards ultimate expert specialization in mixture-of-experts language models},
  author={Dai, Damai and Deng, Chengqi and Zhao, Chenggang and Xu, RX and Gao, Huazuo and Chen, Deli and Li, Jiashi and Zeng, Wangding and Yu, Xingkai and Wu, Yu and others},
  booktitle={Proceedings of the 62nd Annual Meeting of the Association for Computational Linguistics (Volume 1: Long Papers)},
  pages={1280--1297},
  year={2024}
}

@inproceedings{zeng2024adamoe,
  title={Adamoe: Token-adaptive routing with null experts for mixture-of-experts language models},
  author={Zeng, Zihao and Miao, Yibo and Gao, Hongcheng and Zhang, Hao and Deng, Zhijie},
  booktitle={Findings of the Association for Computational Linguistics: EMNLP 2024},
  pages={6223--6235},
  year={2024}
}

@inproceedings{lu2024not,
  title={Not all experts are equal: Efficient expert pruning and skipping for mixture-of-experts large language models},
  author={Lu, Xudong and Liu, Qi and Xu, Yuhui and Zhou, Aojun and Huang, Siyuan and Zhang, Bo and Yan, Junchi and Li, Hongsheng},
  booktitle={Proceedings of the 62nd Annual Meeting of the Association for Computational Linguistics (Volume 1: Long Papers)},
  pages={6159--6172},
  year={2024}
}

@article{jin2024moe++,
  title={Moe++: Accelerating mixture-of-experts methods with zero-computation experts},
  author={Jin, Peng and Zhu, Bo and Yuan, Li and Yan, Shuicheng},
  journal={arXiv preprint arXiv:2410.07348},
  year={2024}
}

@article{team2025longcat,
  title={Longcat-flash technical report},
  author={Team, Meituan LongCat and Li, Bei and Lei, Bingye and Wang, Bo and Rong, Bolin and Wang, Chao and Zhang, Chao and Gao, Chen and Zhang, Chen and Sun, Cheng and others},
  journal={arXiv preprint arXiv:2509.01322},
  year={2025}
}

@article{jiang2024mixtral,
  title={Mixtral of experts},
  author={Jiang, Albert Q and Sablayrolles, Alexandre and Roux, Antoine and Mensch, Arthur and Savary, Blanche and Bamford, Chris and Chaplot, Devendra Singh and Casas, Diego de las and Hanna, Emma Bou and Bressand, Florian and others},
  journal={arXiv preprint arXiv:2401.04088},
  year={2024}
}

@article{liu2024efficient,
  title={Efficient expert pruning for sparse mixture-of-experts language models: Enhancing performance and reducing inference costs},
  author={Liu, Enshu and Zhu, Junyi and Lin, Zinan and Ning, Xuefei and Blaschko, Matthew B and Yan, Shengen and Dai, Guohao and Yang, Huazhong and Wang, Yu},
  journal={arXiv preprint arXiv:2407.00945},
  year={2024}
}

@article{li2023merge,
  title={Merge, then compress: Demystify efficient smoe with hints from its routing policy},
  author={Li, Pingzhi and Zhang, Zhenyu and Yadav, Prateek and Sung, Yi-Lin and Cheng, Yu and Bansal, Mohit and Chen, Tianlong},
  journal={arXiv preprint arXiv:2310.01334},
  year={2023}
}

@inproceedings{chen2025eac,
  title={EAC-MoE: Expert-selection aware compressor for mixture-of-experts large language models},
  author={Chen, Yuanteng and Shao, Yuantian and Wang, Peisong and Cheng, Jian},
  booktitle={Proceedings of the 63rd Annual Meeting of the Association for Computational Linguistics (Volume 1: Long Papers)},
  pages={12942--12963},
  year={2025}
}

@inproceedings{zhang2025diversifying,
  title={Diversifying the expert knowledge for task-agnostic pruning in sparse mixture-of-experts},
  author={Zhang, Zeliang and Liu, Xiaodong and Cheng, Hao and Xu, Chenliang and Gao, Jianfeng},
  booktitle={Findings of the Association for Computational Linguistics: ACL 2025},
  pages={86--102},
  year={2025}
}

@article{hao2026lightmoe,
  title={LightMoE: Reducing Mixture-of-Experts Redundancy through Expert Replacing},
  author={Hao, Jiawei and Hao, Zhiwei and Guo, Jianyuan and Shen, Li and Luo, Yong and Hu, Han and Zeng, Dan},
  journal={arXiv preprint arXiv:2603.12645},
  year={2026}
}

@article{chen2024retraining,
  title={Retraining-free merging of sparse moe via hierarchical clustering},
  author={Chen, I and Liu, Hsu-Shen and Sun, Wei-Fang and Chao, Chen-Hao and Hsu, Yen-Chang and Lee, Chun-Yi and others},
  journal={arXiv preprint arXiv:2410.08589},
  year={2024}
}

@article{zhou2022mixture,
  title={Mixture-of-experts with expert choice routing},
  author={Zhou, Yanqi and Lei, Tao and Liu, Hanxiao and Du, Nan and Huang, Yanping and Zhao, Vincent and Dai, Andrew M and Le, Quoc V and Laudon, James and others},
  journal={Advances in Neural Information Processing Systems},
  volume={35},
  pages={7103--7114},
  year={2022}
}

@inproceedings{huang2024harder,
  title={Harder task needs more experts: Dynamic routing in MoE models},
  author={Huang, Quzhe and An, Zhenwei and Zhuang, Nan and Tao, Mingxu and Zhang, Chen and Jin, Yang and Xu, Kun and Chen, Liwei and Huang, Songfang and Feng, Yansong},
  booktitle={Proceedings of the 62nd Annual Meeting of the Association for Computational Linguistics (Volume 1: Long Papers)},
  pages={12883--12895},
  year={2024}
}

@inproceedings{yue2024ada,
  title={Ada-k routing: Boosting the efficiency of moe-based llms},
  author={Yue, Tongtian and Guo, Longteng and Cheng, Jie and Gao, Xuange and Huang, Hua and Liu, Jing},
  booktitle={The Thirteenth International Conference on Learning Representations},
  year={2024}
}

@article{guo2024dynamic,
  title={Dynamic mixture of experts: An auto-tuning approach for efficient transformer models},
  author={Guo, Yongxin and Cheng, Zhenglin and Tang, Xiaoying and Tu, Zhaopeng and Lin, Tao},
  journal={arXiv preprint arXiv:2405.14297},
  year={2024}
}

@article{sun2026expert,
  title={Expert Threshold Routing for Autoregressive Language Modeling with Dynamic Computation Allocation and Load Balancing},
  author={Sun, Hanchi and Liu, Yixin and Wu, Yonghui and Sun, Lichao},
  journal={arXiv preprint arXiv:2603.11535},
  year={2026}
}

@article{hinton2015distilling,
  title={Distilling the knowledge in a neural network},
  author={Hinton, Geoffrey and Vinyals, Oriol and Dean, Jeff},
  journal={arXiv preprint arXiv:1503.02531},
  year={2015}
}

@article{gu2023minillm,
  title={Minillm: Knowledge distillation of large language models},
  author={Gu, Yuxian and Dong, Li and Wei, Furu and Huang, Minlie},
  journal={arXiv preprint arXiv:2306.08543},
  year={2023}
}

@inproceedings{agarwal2024policy,
  title={On-policy distillation of language models: Learning from self-generated mistakes},
  author={Agarwal, Rishabh and Vieillard, Nino and Zhou, Yongchao and Stanczyk, Piotr and Garea, Sabela Ramos and Geist, Matthieu and Bachem, Olivier},
  booktitle={The twelfth international conference on learning representations},
  year={2024}
}

@inproceedings{furlanello2018born,
  title={Born again neural networks},
  author={Furlanello, Tommaso and Lipton, Zachary and Tschannen, Michael and Itti, Laurent and Anandkumar, Anima},
  booktitle={International conference on machine learning},
  pages={1607--1616},
  year={2018},
  organization={PMLR}
}

@inproceedings{zhang2019your,
  title={Be your own teacher: Improve the performance of convolutional neural networks via self distillation},
  author={Zhang, Linfeng and Song, Jiebo and Gao, Anni and Chen, Jingwei and Bao, Chenglong and Ma, Kaisheng},
  booktitle={Proceedings of the IEEE/CVF international conference on computer vision},
  pages={3713--3722},
  year={2019}
}

@article{zhao2026self,
  title={Self-Distilled Reasoner: On-Policy Self-Distillation for Large Language Models},
  author={Zhao, Siyan and Xie, Zhihui and Liu, Mengchen and Huang, Jing and Pang, Guan and Chen, Feiyu and Grover, Aditya},
  journal={arXiv preprint arXiv:2601.18734},
  year={2026}
}

@article{shenfeld2026self,
  title={Self-Distillation Enables Continual Learning},
  author={Shenfeld, Idan and Damani, Mehul and H{\"u}botter, Jonas and Agrawal, Pulkit},
  journal={arXiv preprint arXiv:2601.19897},
  year={2026}
}

@article{sanh2019distilbert,
  title={DistilBERT, a distilled version of BERT: smaller, faster, cheaper and lighter},
  author={Sanh, Victor and Debut, Lysandre and Chaumond, Julien and Wolf, Thomas},
  journal={arXiv preprint arXiv:1910.01108},
  year={2019}
}

@inproceedings{kim2016sequence,
  title={Sequence-level knowledge distillation},
  author={Kim, Yoon and Rush, Alexander M},
  booktitle={Proceedings of the 2016 conference on empirical methods in natural language processing},
  pages={1317--1327},
  year={2016}
}

@inproceedings{hsieh2023distilling,
  title={Distilling step-by-step! outperforming larger language models with less training data and smaller model sizes},
  author={Hsieh, Cheng-Yu and Li, Chun-Liang and Yeh, Chih-Kuan and Nakhost, Hootan and Fujii, Yasuhisa and Ratner, Alex and Krishna, Ranjay and Lee, Chen-Yu and Pfister, Tomas},
  booktitle={Findings of the Association for Computational Linguistics: ACL 2023},
  pages={8003--8017},
  year={2023}
}

@article{lu2025onpolicydistillation,
  author = {Kevin Lu and Thinking Machines Lab},
  title = {On-Policy Distillation},
  journal = {Thinking Machines Lab: Connectionism},
  year = {2025},
  note = {https://thinkingmachines.ai/blog/on-policy-distillation},
  doi = {10.64434/tml.20251026},
}

@article{hoshino2025rad,
  title={RAD: Redundancy-Aware Distillation for Hybrid Models via Self-Speculative Decoding},
  author={Hoshino, Yuichiro and Tachibana, Hideyuki and Inahara, Muneyoshi and Takegawa, Hiroto},
  journal={arXiv preprint arXiv:2505.22135},
  year={2025}
}

@article{chen2026hybrid,
  title={Hybrid Linear Attention Done Right: Efficient Distillation and Effective Architectures for Extremely Long Contexts},
  author={Chen, Yingfa and Thai, Zhen Leng and Zhou, Zihan and Zhang, Zhu and Shen, Xingyu and Wang, Shuo and Xiao, Chaojun and Han, Xu and Liu, Zhiyuan},
  journal={arXiv preprint arXiv:2601.22156},
  year={2026}
}

@article{kim2024ladimo,
  title={Ladimo: Layer-wise distillation inspired moefier},
  author={Kim, Sungyoon and Kim, Youngjun and Moon, Kihyo and Jang, Minsung},
  journal={arXiv preprint arXiv:2408.04278},
  year={2024}
}

@article{hubotter2026reinforcement,
  title={Reinforcement Learning via Self-Distillation},
  author={H{\"u}botter, Jonas and L{\"u}beck, Frederike and Behric, Lejs and Baumann, Anton and Bagatella, Marco and Marta, Daniel and Hakimi, Ido and Shenfeld, Idan and Buening, Thomas Kleine and Guestrin, Carlos and others},
  journal={arXiv preprint arXiv:2601.20802},
  year={2026}
}

@inproceedings{ainslie2023gqa,
  title={Gqa: Training generalized multi-query transformer models from multi-head checkpoints},
  author={Ainslie, Joshua and Lee-Thorp, James and De Jong, Michiel and Zemlyanskiy, Yury and Lebr{\'o}n, Federico and Sanghai, Sumit},
  booktitle={Proceedings of the 2023 Conference on Empirical Methods in Natural Language Processing},
  pages={4895--4901},
  year={2023}
}

@article{yang2025qwen3,
  title={Qwen3 technical report},
  author={Yang, An and Li, Anfeng and Yang, Baosong and Zhang, Beichen and Hui, Binyuan and Zheng, Bo and Yu, Bowen and Gao, Chang and Huang, Chengen and Lv, Chenxu and others},
  journal={arXiv preprint arXiv:2505.09388},
  year={2025}
}

@article{zheng2024sglang,
  title={Sglang: Efficient execution of structured language model programs},
  author={Zheng, Lianmin and Yin, Liangsheng and Xie, Zhiqiang and Sun, Chuyue and Huang, Jeff and Yu, Cody H and Cao, Shiyi and Kozyrakis, Christos and Stoica, Ion and Gonzalez, Joseph E and others},
  journal={Advances in neural information processing systems},
  volume={37},
  pages={62557--62583},
  year={2024}
}

@article{lepikhin2020gshard,
  title={Gshard: Scaling giant models with conditional computation and automatic sharding},
  author={Lepikhin, Dmitry and Lee, HyoukJoong and Xu, Yuanzhong and Chen, Dehao and Firat, Orhan and Huang, Yanping and Krikun, Maxim and Shazeer, Noam and Chen, Zhifeng},
  journal={arXiv preprint arXiv:2006.16668},
  year={2020}
}

@article{zeng2025glm,
  title={Glm-4.5: Agentic, reasoning, and coding (arc) foundation models},
  author={Zeng, Aohan and Lv, Xin and Zheng, Qinkai and Hou, Zhenyu and Chen, Bin and Xie, Chengxing and Wang, Cunxiang and Yin, Da and Zeng, Hao and Zhang, Jiajie and others},
  journal={arXiv preprint arXiv:2508.06471},
  year={2025}
}

@article{cobbe2021training,
  title={Training verifiers to solve math word problems},
  author={Cobbe, Karl and Kosaraju, Vineet and Bavarian, Mohammad and Chen, Mark and Jun, Heewoo and Kaiser, Lukasz and Plappert, Matthias and Tworek, Jerry and Hilton, Jacob and Nakano, Reiichiro and others},
  journal={arXiv preprint arXiv:2110.14168},
  year={2021}
}

@inproceedings{lightman2023let,
  title={Let's verify step by step},
  author={Lightman, Hunter and Kosaraju, Vineet and Burda, Yuri and Edwards, Harrison and Baker, Bowen and Lee, Teddy and Leike, Jan and Schulman, John and Sutskever, Ilya and Cobbe, Karl},
  booktitle={The twelfth international conference on learning representations},
  year={2023}
}

@article{liu2023your,
  title={Is your code generated by chatgpt really correct? rigorous evaluation of large language models for code generation},
  author={Liu, Jiawei and Xia, Chunqiu Steven and Wang, Yuyao and Zhang, Lingming},
  journal={Advances in neural information processing systems},
  volume={36},
  pages={21558--21572},
  year={2023}
}

@article{zhou2023instruction,
  title={Instruction-following evaluation for large language models},
  author={Zhou, Jeffrey and Lu, Tianjian and Mishra, Swaroop and Brahma, Siddhartha and Basu, Sujoy and Luan, Yi and Zhou, Denny and Hou, Le},
  journal={arXiv preprint arXiv:2311.07911},
  year={2023}
}

@article{pyatkin2025generalizing,
  title={Generalizing verifiable instruction following},
  author={Pyatkin, Valentina and Malik, Saumya and Graf, Victoria and Ivison, Hamish and Huang, Shengyi and Dasigi, Pradeep and Lambert, Nathan and Hajishirzi, Hannaneh},
  journal={arXiv preprint arXiv:2507.02833},
  year={2025}
}

@inproceedings{gema2025we,
  title={Are we done with mmlu?},
  author={Gema, Aryo Pradipta and Leang, Joshua Ong Jun and Hong, Giwon and Devoto, Alessio and Mancino, Alberto Carlo Maria and Saxena, Rohit and He, Xuanli and Zhao, Yu and Du, Xiaotang and Madani, Mohammad Reza Ghasemi and others},
  booktitle={Proceedings of the 2025 Conference of the Nations of the Americas Chapter of the Association for Computational Linguistics: Human Language Technologies (Volume 1: Long Papers)},
  pages={5069--5096},
  year={2025}
}

@article{rein2023gpqa,
  title={Gpqa: A graduate-level google-proof q\&a benchmark},
  author={Rein, David and Hou, Betty Li and Stickland, Asa Cooper and Petty, Jackson and Pang, Richard Yuanzhe and Dirani, Julien and Michael, Julian and Bowman, Samuel R},
  journal={arXiv preprint arXiv:2311.12022},
  year={2023}
}

@article{ko2026scaling,
  title={Scaling reasoning efficiently via relaxed on-policy distillation},
  author={Ko, Jongwoo and Abdali, Sara and Kim, Young Jin and Chen, Tianyi and Cameron, Pashmina},
  journal={arXiv preprint arXiv:2603.11137},
  year={2026}
}

@misc{deepep2025,
      title={DeepEP: an efficient expert-parallel communication library},
      author={Chenggang Zhao and Shangyan Zhou and Liyue Zhang and Chengqi Deng and Zhean Xu and Yuxuan Liu and Kuai Yu and Jiashi Li and Liang Zhao},
      year={2025},
      publisher = {GitHub},
      howpublished = {https://github.com/deepseek-ai/DeepEP},
}

@article{shoeybi2019megatron,
  title={Megatron-lm: Training multi-billion parameter language models using model parallelism},
  author={Shoeybi, Mohammad and Patwary, Mostofa and Puri, Raul and LeGresley, Patrick and Casper, Jared and Catanzaro, Bryan},
  journal={arXiv preprint arXiv:1909.08053},
  year={2019}
}

@misc{slime_github,
  author       = {Zilin Zhu and Chengxing Xie and Xin Lv and slime Contributors},
  title        = {slime: An LLM post-training framework for RL Scaling},
  year         = {2025},
  howpublished = {https://github.com/THUDM/slime},
  note         = {GitHub repository. Corresponding author: Xin Lv},
  urldate      = {2025-06-19}
}

@misc{qwen35blog,
    title = {Qwen3.5: Accelerating Productivity with Native Multimodal Agents},
    author = {Qwen Team},
    month = {February},
    year = {2026}
}

@article{zeng2026glm5,
  title={Glm-5: from vibe coding to agentic engineering},
  author={Zeng, Aohan and Lv, Xin and Hou, Zhenyu and Du, Zhengxiao and Zheng, Qinkai and Chen, Bin and Yin, Da and Ge, Chendi and Huang, Chenghua and Xie, Chengxing and others},
  journal={arXiv preprint arXiv:2602.15763},
  year={2026}
}

@misc{deepseekai2026deepseekv4,
      title={DeepSeek-V4: Towards Highly Efficient Million-Token Context Intelligence},
      author={DeepSeek-AI},
      year={2026},
}

@article{liu2025acereason,
  title={Acereason-nemotron 1.1: Advancing math and code reasoning through sft and rl synergy},
  author={Liu, Zihan and Yang, Zhuolin and Chen, Yang and Lee, Chankyu and Shoeybi, Mohammad and Catanzaro, Bryan and Ping, Wei},
  journal={arXiv preprint arXiv:2506.13284},
  year={2025}
}

@article{bercovich2025llama,
  title={Llama-nemotron: Efficient reasoning models},
  author={Bercovich, Akhiad and Levy, Itay and Golan, Izik and Dabbah, Mohammad and El-Yaniv, Ran and Puny, Omri and Galil, Ido and Moshe, Zach and Ronen, Tomer and Nabwani, Najeeb and others},
  journal={arXiv preprint arXiv:2505.00949},
  year={2025}
}

@article{wu2025grove,
  title={Grove moe: Towards efficient and superior moe llms with adjugate experts},
  author={Wu, Haoyuan and Chen, Haoxing and Chen, Xiaodong and Zhou, Zhanchao and Chen, Tieyuan and Zhuang, Yihong and Lu, Guoshan and Huang, Zenan and Zhao, Junbo and Liu, Lin and others},
  journal={arXiv preprint arXiv:2508.07785},
  year={2025}
}

@article{chaudhari2026moe,
  title={MoE Lens--An Expert Is All You Need},
  author={Chaudhari, Marmik and Gulati, Idhant and Hundia, Nishkal and Karra, Pranav and Raval, Shivam},
  journal={arXiv preprint arXiv:2603.05806},
  year={2026}
}

\appendix
\newpage

\section{Limitations and Future Work}
\label{app:limitations}

\paragraph{Lack of Larger MoE Deployments.} Although we demonstrate consistent improvements on 30B-scale MoE models, we do not yet evaluate substantially larger-scale MoE models due to computational resource constraints.

\paragraph{Lack of Long-Horizon Agentic Tasks.} Our experimental evaluation is confined to standard post-training tasks and does not cover agentic workloads. A contributing factor is the limited availability of mature open-source agentic infrastructure and training recipes.

\paragraph{Speedup Decay at Long Sequence Lengths.} 
Beyond the $8k$ results reported in the main text, we also evaluate sequence lengths of \(\{2k, 4k, 6k, 8k\}\), as shown in Table~\ref{tab:inference-efficiency-limitation}. The speedup gradually diminishes as sequence length increases. Nevertheless, even at $8k$, a commonly used long-context setting, ZEDA still achieves approximately 20\% speedup, demonstrating its practical usability. Furthermore, ZEDA exhibits greater potential for advanced communication frameworks like DeepEP \citep{deepep2025}, which we aim to integrate in future work.

\begin{table*}[h]
  \centering
  \caption{Inference efficiency comparison between the original model and ZEDA from 2048 to 8192 sequence length. Speedup is defined relative to the original model and throughput ($10^3$ token$\cdot$s$^{-1}$) is shown in the format ZEDA/original.}
  \label{tab:inference-efficiency-limitation}
  \resizebox{\textwidth}{!}{%
  \begin{tabular}{llcccc|cccc}
  \toprule
  \multirow{2}{*}{\textbf{Model}} & \multirow{2}{*}{\textbf{Metric}} & \multicolumn{4}{c|}{\textbf{Prefill}} & \multicolumn{4}{c}{\textbf{Decode}} \\
  \cmidrule(lr){3-6} \cmidrule(lr){7-10}
   &  & \textbf{2048} & \textbf{4096} & \textbf{6144} & \textbf{8192} & \textbf{2048} & \textbf{4096} & \textbf{6144} & \textbf{8192} \\
  \midrule
  \multirow{2}{*}{\textbf{Qwen3-30B-A3B}} & \cellcolor{lightblue!100} Speedup & \cellcolor{lightblue!100}1.21x & \cellcolor{lightblue!100}1.20x & \cellcolor{lightblue!100}1.19x & \cellcolor{lightblue!100}1.18x & \cellcolor{lightblue!100}1.25x & \cellcolor{lightblue!100}1.24x & \cellcolor{lightblue!100}1.21x & \cellcolor{lightblue!100}1.19x \\
   & Throughput & 71.43/58.86 & 62.53/51.97 & 58.09/48.75 & 51.63/43.92 & 2.49/1.99 & 2.39/1.93 & 2.22/1.82 & 2.07/1.74 \\
  \multirow{2}{*}{\textbf{GLM-4.7-Flash}} & \cellcolor{lightblue!100} Speedup & \cellcolor{lightblue!100}1.36x & \cellcolor{lightblue!100}1.31x & \cellcolor{lightblue!100}1.27x & \cellcolor{lightblue!100}1.26x & \cellcolor{lightblue!100}1.22x & \cellcolor{lightblue!100}1.20x & \cellcolor{lightblue!100}1.19x & \cellcolor{lightblue!100}1.19x \\
   & Throughput & 47.78/35.11 & 39.42/30.20 & 34.37/27.06 & 32.36/25.73 & 1.91/1.57 & 1.88/1.56 & 1.80/1.51 & 1.76/1.48 \\
  \bottomrule
  \end{tabular}
  }
  \end{table*}

\section{Zero Experts versus Copy Experts}
\label{sec:zero_vs_copy}

An alternative to the zero expert is the \emph{copy expert}, whose output equals its input, carrying negligible computational cost. In this section, we establish that zero expert is the preferable design for adapting a post-trained MoE model to a dynamic architecture.

Compared with zero experts, copy experts introduce stronger perturbations to the original model. Assuming the same router parameterization, the output of a dynamic MoE module with zero experts and that with copy experts is 
\begin{equation}
\tilde{y}_{\text{zero}} = \sum_{i \in \tilde{\mathcal{S}} \cap \mathcal{E}}\tilde{g}_i(h)\, E_i(h), \quad
\tilde{y}_{\text{copy}} = \tilde{y}_{\text{copy}}^{\text{norm}} + \tilde{y}_{\text{copy}}^{\text{cp}}
= \sum_{i \in \tilde{\mathcal{S}} \cap \mathcal{E}}
\tilde{g}_i(h)\, E_i(h) + \sum_{j \in \tilde{\mathcal{S}} \cap \mathcal{Z}} \tilde{g}_j(h)\, h,
\end{equation}
where $\tilde{y}_{\text{copy}}^{\text{norm}}$ and $\tilde{y}_{\text{copy}}^{\text{cp}}$ are the normal expert component and copy component of the copy-expert model, respectively.
Zero experts implement true expert omission, whereas copy experts incur an additional term $\sum_{j \in \tilde{\mathcal{S}}(h) \cap \mathcal{Z}} \tilde{g}_j(h)\, h$ rather than a no-op.

To study the effect of the zero-compute-expert type, we compare two post-training adaptation variants on Qwen3-30B-A3B that are identical in training data, SFT recipe, and routing regularization, differing only in whether the inserted zero-compute experts are instantiated as copy experts or zero experts. For both variants, the routing regularizer is the group auxiliary loss $\mathcal{L}_{GA}$ in Eq.~\eqref{eq:gal} with coefficient $\alpha{=}0.1$ and group weight $w{=}2.0$, matching the setting used in Section~\ref{sec:settings}. We report the average accuracy, the average activation ratio of the inserted zero-compute experts $r_{ZCE}$, and the performance on five mathematical reasoning benchmarks. Table~\ref{tab:copy-zero-math} shows that the copy-expert type performs substantially worse than the zero-expert type despite nearly identical activation ratios ($53.2\%$ vs.\ $52.7\%$). While the zero-expert type largely preserves the mathematical reasoning ability of the original model, the copy-expert type leads to severe performance degradation across all five benchmarks. The gap is particularly pronounced on the more challenging AIME tasks, where the copy-expert type achieves only $1.0$, $2.9$, and $0.8$ on AIME 24, AIME 25, and AIME 26, respectively. These empirical results suggest that, for post-training adaptation of a post-trained MoE model toward dynamic computation, the zero-expert type is substantially more suitable than the copy-expert type.

\begin{table*}[t]
\definecolor{mygray}{rgb}{0.85,0.85,0.85}
\centering
\caption{Comparison of two zero-compute-expert types, copy expert and zero expert, on mathematical reasoning benchmarks.}
\label{tab:copy-zero-math}
\resizebox{0.8\textwidth}{!}{
\begin{tabular}{lcc|ccccc}
\toprule
\multirow{2}{*}[-3pt]{\textbf{Method}} & \multirow{2}{*}[-3pt]{\shortstack{\textbf{Avg}\\\textbf{Acc}}} & \multirow{2}{*}[-3pt]{\shortstack{\textbf{Avg}\\\textbf{$r_{ZCE}$}}} & \multicolumn{5}{c}{\textbf{Math}} \\
\cmidrule(lr){4-8}
 &  &  & \textbf{AIME 24} & \textbf{AIME 25} & \textbf{AIME 26} & \textbf{GSM8K} & \textbf{MATH-500} \\
\midrule
\rowcolor{mygray!100} \textbf{Qwen3-30B-A3B} & $82.8$ & $0.0$ & $80.9$ & $71.0$ & $72.3$ & $95.4$ & $94.4$ \\
Copy Expert + $\mathcal{L}_{GA}$ & $20.7$ & $53.2$ & $1.0$ & $2.9$ & $0.8$ & $58.8$ & $40.0$ \\
\rowcolor{lightblue!100} Zero Expert + $\mathcal{L}_{GA}$ & $81.0$ & $52.7$ & $78.1$ & $66.2$ & $71.3$ & $94.8$ & $94.4$ \\
\bottomrule
\end{tabular}}
\end{table*}

We extract the hidden-state outputs of the MoE blocks and conduct the following experiments to further analyze why copy experts are harmful from two complementary perspectives:
\begin{figure*}[t]
  \centering
  \includegraphics[width=0.95\textwidth]{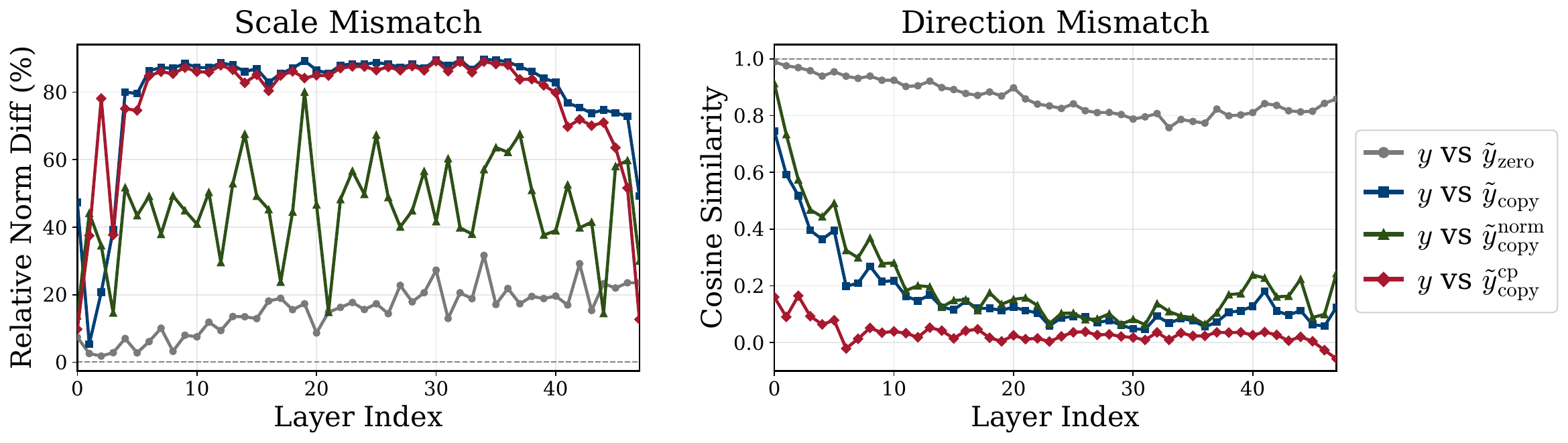}
  \caption{Layer-wise comparison between the original MoE output $y$ and four variants: $\tilde{y}_{\text{zero}}$, $\tilde{y}_{\text{copy}}$, $\tilde{y}_{\text{copy}}^{\text{norm}}$ and $\tilde{y}_{\text{copy}}^{\text{cp}}$, including relative absolute L2-norm difference (left) and cosine similarity (right).}
  \label{fig:zero_vs_copy_mismatch}
\end{figure*}
\begin{itemize}[leftmargin=1em]

\item \textbf{Scale Mismatch.} The left panel of Figure~\ref{fig:zero_vs_copy_mismatch} shows that the full copy expert output has a much larger norm mismatch with the original output than the zero expert output does.
Decomposition further shows that the normal expert component is less mismatched than the copy component, but still consistently more mismatched than the zero-expert output. This indicates that the copy component is the main source of scale mismatch, and also pulls the normal expert component away from the original scale.

\item \textbf{Direction Mismatch.} The right panel of Figure~\ref{fig:zero_vs_copy_mismatch} shows the same trend in direction space. 
The zero expert output remains well aligned with the original output, whereas the full copy expert output shows a clear directional mismatch. The copy component stays strongly misaligned across layers, while the mismatch of the normal expert component grows from shallow to deep layers. This suggests that the copy component is the primary cause of directional mismatch and progressively drags the normal expert component away from the original direction as depth increases.

\end{itemize}

\section{Auxiliary-Loss Comparison}
\label{app:aux-loss-comparison}

\paragraph{Experimental Setting.}
To isolate the effect of the routing regularizer in Section~\ref{sec:gal}, we compare three zero-expert adaptation variants on Qwen3-30B-A3B that are identical in architecture and SFT adaptation recipe, differing only in the balancing objective: one uses the standard expert-level auxiliary loss $\mathcal{L}_A$ in Eq.~\eqref{eq:aux}, and the other two use the proposed group auxiliary loss $\mathcal{L}_{GA}$ in Eq.~\eqref{eq:gal}. The experimental setting is otherwise the same as the SFT implementation details in Section~\ref{sec:settings}. In particular, the auxiliary-loss coefficient $\alpha$ is set to 0.1 for all variants. For the group auxiliary loss, the relative weight of the zero-expert group $w$ is set to 1.0 and 2.0, respectively. We report the results on five mathematical reasoning benchmarks.

\begin{table*}[h]
\definecolor{mygray}{rgb}{0.85,0.85,0.85}
\centering
\caption{Comparison of auxiliary loss and group auxiliary loss under different $w$ values on mathematical reasoning benchmarks.}
\label{tab:aux-loss-math}
\resizebox{0.8\textwidth}{!}{
\begin{tabular}{lcc|ccccc}
\toprule
\multirow{2}{*}[-3pt]{\textbf{Method}} & \multirow{2}{*}[-3pt]{\shortstack{\textbf{Avg}\\\textbf{Acc}}} & \multirow{2}{*}[-3pt]{\shortstack{\textbf{Avg}\\\textbf{$r_{ZE}$}}} & \multicolumn{5}{c}{\textbf{Math}} \\
\cmidrule(lr){4-8}
 &  &  & \textbf{AIME 24} & \textbf{AIME 25} & \textbf{AIME 26} & \textbf{GSM8K} & \textbf{MATH-500} \\
\midrule
\rowcolor{mygray!100} \textbf{Qwen3-30B-A3B} & $82.8$ & $0.0$ & $80.9$ & $71.0$ & $72.3$ & $95.4$ & $94.4$ \\
Zero Expert + $\mathcal{L}_{A}$ & $59.5$ & $34.4$ & $39.6$ & $32.3$ & $47.1$ & $93.5$ & $85.0$ \\
\rowcolor{lightblue!100} Zero Expert + $\mathcal{L}_{GA}$ $(w{=}1.0)$ & $82.2$ & $35.0$ & $79.4$ & $69.7$ & $71.9$ & $95.2$ & $94.8$ \\
Zero Expert + $\mathcal{L}_{GA}$ $(w{=}2.0)$ & $81.0$ & $52.7$ & $78.1$ & $66.2$ & $71.3$ & $94.8$ & $94.4$ \\
\bottomrule
\end{tabular}}
\end{table*}

\paragraph{Conclusion.}
Compared with the original auxiliary loss, both group auxiliary loss variants deliver significant improvements, showing that the benefit of group-level balancing is robust across different $w$ settings. In particular, replacing $\mathcal{L}_{A}$ with $\mathcal{L}_{GA}$ at $w{=}1.0$ improves average accuracy from $59.5$ to $82.2$, nearly recovering the original Qwen3-30B-A3B performance of $82.8$, while maintaining a similar zero-expert activation ratio. Increasing the group weight to $w{=}2.0$ raises $r_{ZE}$ further to $52.7$ while preserving a strong average accuracy of $81.0$, indicating a clear quality-efficiency trade-off.

These results are consistent with the design motivation in Section~\ref{sec:gal}. A post-trained MoE model exhibits non-uniform, input-dependent routing patterns over normal experts, and enforcing expert-level uniformity disrupts these learned routing distributions, which can severely degrade model performance. By contrast, the group auxiliary loss regulates only the competition between the normal-expert group and the zero-expert group, thereby preserving the relative routing structure among normal experts while enabling controllable zero-expert utilization through $w$.

\section{Theoretical FLOPs Analysis}
\label{app:flop}

In this section, we analyze the theoretical FLOPs of both the prefill and decode stages for the original MoE model and the model adapted by ZEDA. We focus on dominant matrix multiplication terms and omit lower-order operations
such as normalization, residual connections, activation functions, routing top-$k$ selection, and softmax overhead. For a matrix multiplication \([m,n] \times [n,p]\), we count its cost as \(2mnp\) FLOPs. All expressions are reported per Transformer layer. Multiplying by the number of layers does not change the ZEDA/original FLOP ratios when all layers share the same configuration.

The notation used throughout this section is summarized in Table~\ref{tab:flop-notation}.

\begin{table}[h]
\centering
\caption{Notation used in the theoretical FLOP analysis.}
\label{tab:flop-notation}
\begin{tabular}{ll}
\toprule
Symbol & Description \\
\midrule
\(l\) & Sequence length in the stage under analysis \\
\(H\) & Hidden size \\
\(H_{\mathrm{attn}}\) & Attention intermediate size \\
\(g_{\mathrm{kv}}\) & Ratio of KV heads to query heads in GQA \\
\(H_e\) & Expert intermediate size \\
\(N\) & Number of normal experts \\
\(N_Z\) & Number of zero-computation experts \\
\(K\) & Number of activated experts per token \\
\(r_{ZE}\) & Fraction of activated zero experts \\
\bottomrule
\end{tabular}
\end{table}

\subsection{Shared MoE Cost Decomposition}

The MoE FFN and router costs have the same form in both stages; the only difference is the number of tokens processed in the current forward pass. Let \(n\) denote that token count. For the original MoE model, each token activates \(K\) normal experts, and each expert contains up, gate, and down projections. Hence, the expert FFN and router costs are

\begin{equation}\label{eq:moe-flop-origin}
F_{\mathrm{MoE,orig}}(n) = 6KnHH_e + 2NnH.
\end{equation}

For the ZEDA model, only an \((1-r_{ZE})\) fraction of the activated experts perform FFN computation, while the router scores both normal and zero-computation experts. Therefore,
\begin{equation}\label{eq:moe-flop-zeda}
F_{\mathrm{MoE,ZEDA}}(n) = 6(1-r_{ZE}) KnHH_e + 2(N+N_Z)nH.
\end{equation}

In the prefill stage, \(n=l\). In the decode stage with KV cache, each forward pass processes one newly generated token, and the total decode cost is obtained by summing over all decode steps.

\subsection{Prefill Stage}

For grouped-query attention (GQA)~\citep{ainslie2023gqa}, the prefill stage processes all \(l\) tokens in parallel. The attention cost therefore consists of five parts: the query projection, the key/value projections, the query-key score computation over all token pairs, the attention-value aggregation, and the output projection. These terms sum to
\begin{equation}\label{eq:attn-flop-prefill}
F_{\mathrm{attn}}^{\mathrm{pre}}
=
4l^2H_{\mathrm{attn}}
+
4(1+g_{\mathrm{kv}})lHH_{\mathrm{attn}}.
\end{equation}

Substituting \(n=l\) into Equations~\eqref{eq:moe-flop-origin} and~\eqref{eq:moe-flop-zeda}, and then adding the prefill attention term in Equation~\eqref{eq:attn-flop-prefill}, yields the total prefill FLOPs of the original model and the ZEDA model. In both expressions, the first two terms come from attention, the third term is the MoE FFN cost, and the last term is the router cost:
\begin{equation}\label{eq:total-flop-prefill-origin}
F_{\mathrm{orig}}^{\mathrm{pre}}
=
4l^2H_{\mathrm{attn}}
+
4(1+g_{\mathrm{kv}})lHH_{\mathrm{attn}}
+
6KlHH_e
+
2NlH.
\end{equation}

For the ZEDA model, the attention term remains unchanged, while the expert FFN cost is reduced by the factor \((1-r_{ZE})\) and the router cost increases because the router now scores \(N+N_Z\) experts:
\begin{equation}\label{eq:total-flop-prefill-zeda}
F_{\mathrm{ZEDA}}^{\mathrm{pre}}
=
4l^2H_{\mathrm{attn}}
+
4(1+g_{\mathrm{kv}})lHH_{\mathrm{attn}}
+
6(1-r_{ZE}) KlHH_e
+
2(N+N_Z)lH.
\end{equation}

The corresponding FLOP ratio, obtained from Equations~\eqref{eq:total-flop-prefill-origin} and~\eqref{eq:total-flop-prefill-zeda}, is
\begin{equation}\label{eq:flop-ratio-prefill}
\boxed{
\frac{F_{\mathrm{ZEDA}}^{\mathrm{pre}}}{F_{\mathrm{orig}}^{\mathrm{pre}}}
=
\frac{
2lH_{\mathrm{attn}}
+
2(1+g_{\mathrm{kv}})HH_{\mathrm{attn}}
+
3(1-r_{ZE}) KHH_e
+
(N+N_Z)H
}{
2lH_{\mathrm{attn}}
+
2(1+g_{\mathrm{kv}})HH_{\mathrm{attn}}
+
3KHH_e
+
NH
}
}
\end{equation}

\subsection{Decode Stage}

In the decode stage, we assume standard KV caching and analyze a decode-only process that generates \(l\) tokens. As in the prefill case, the attention cost consists of query projection, key/value projections, score computation, attention-value aggregation, and output projection. The difference is that at decode step \(t\), only one new token is processed, and the score computation and attention-value aggregation each involve \(t-1\) cached tokens rather than all \(l\) tokens. Summing these per-step costs over all \(l\) decode steps gives
\begin{equation}\label{eq:attn-flop-decode}
\begin{aligned}
F_{\mathrm{attn}}^{\mathrm{dec}}
&=
\sum_{t=1}^{l}
\left[
4(t-1)H_{\mathrm{attn}}
+
4(1+g_{\mathrm{kv}})HH_{\mathrm{attn}}
\right] \\
&=
2l(l-1)H_{\mathrm{attn}}
+
4(1+g_{\mathrm{kv}})lHH_{\mathrm{attn}}.
\end{aligned}
\end{equation}

Substituting the per-token MoE costs from Equations~\eqref{eq:moe-flop-origin} and~\eqref{eq:moe-flop-zeda} across all \(l\) decode steps, and adding the accumulated attention cost in Equation~\eqref{eq:attn-flop-decode}, gives the total decode FLOPs. As in the prefill case, the first two terms correspond to attention, the third term is the MoE FFN cost, and the last term is the router cost:
\begin{equation}\label{eq:total-flop-decode-origin}
F_{\mathrm{orig}}^{\mathrm{dec}}
=
2l(l-1)H_{\mathrm{attn}}
+
4(1+g_{\mathrm{kv}})lHH_{\mathrm{attn}}
+
6KlHH_e
+
2NlH.
\end{equation}

For the ZEDA model, the decode attention term is again identical to that of the original model, whereas the MoE branch differs in exactly the same way as in prefill: only an \((1-r_{ZE})\) fraction of activated experts incur FFN cost, and the router expands from \(N\) to \(N+N_Z\) outputs:
\begin{equation}\label{eq:total-flop-decode-zeda}
F_{\mathrm{ZEDA}}^{\mathrm{dec}}
=
2l(l-1)H_{\mathrm{attn}}
+
4(1+g_{\mathrm{kv}})lHH_{\mathrm{attn}}
+
6(1-r_{ZE}) KlHH_e
+
2(N+N_Z)lH.
\end{equation}

Therefore, the decode-stage FLOP ratio, obtained from Equations~\eqref{eq:total-flop-decode-origin} and~\eqref{eq:total-flop-decode-zeda}, is
\begin{equation}\label{eq:flop-ratio-decode}
\boxed{
\frac{F_{\mathrm{ZEDA}}^{\mathrm{dec}}}{F_{\mathrm{orig}}^{\mathrm{dec}}}
=
\frac{
(l-1)H_{\mathrm{attn}}
+
2(1+g_{\mathrm{kv}})HH_{\mathrm{attn}}
+
3(1-r_{ZE}) KHH_e
+
(N+N_Z)H
}{
(l-1)H_{\mathrm{attn}}
+
2(1+g_{\mathrm{kv}})HH_{\mathrm{attn}}
+
3KHH_e
+
NH
}
}
\end{equation}

\subsection{Numerical Results}

We instantiate the prefill and decode ratios in Equations~\eqref{eq:flop-ratio-prefill} and~\eqref{eq:flop-ratio-decode} using the Qwen3-30B-A3B configuration~\citep{yang2025qwen3} in Table~\ref{tab:flop-qwen-values}.

\begin{table}[h]
\centering
\caption{Architectural parameters of Qwen3-30B-A3B used in the FLOP analysis.}
\label{tab:flop-qwen-values}
\begin{tabular}{lccccccc}
\toprule
Symbol & \(H\) & \(H_{\mathrm{attn}}\) & \(g_{\mathrm{kv}}\) & \(H_e\) & \(N\) & \(N_Z\) & \(K\) \\
\midrule
Value & \(2048\) & \(4096\) & \(1/8\) & \(768\) & \(128\) & \(64\) & \(8\) \\
\bottomrule
\end{tabular}
\end{table}

To facilitate direct comparison with empirical measurements, we convert the FLOP ratios in Equations~\eqref{eq:flop-ratio-prefill} and~\eqref{eq:flop-ratio-decode} into theoretical speedups by taking their reciprocals. Table~\ref{tab:flop-speedup-comparison} reports the resulting prefill and decode speedups for \(l \in \{1024, 2048, \dots, 8192\}\) and \(r_{ZE} = 0.5\), together with the corresponding
empirically measured results.

\begin{table}[h]
  \centering
  \caption{Comparison between theoretical speedups derived from the FLOP analysis and measured empirical speedups on Qwen3-30B-A3B across different sequence lengths.}
  \label{tab:flop-speedup-comparison}
  \resizebox{0.55\textwidth}{!}{
  \begin{tabular}{ccccc}
  \toprule
  \multirow{2}{*}{Length} & \multicolumn{2}{c}{Prefill Speedup} & \multicolumn{2}{c}{Decode Speedup} \\
  \cmidrule(lr){2-3} \cmidrule(lr){4-5}
   & Theoretical & Empirical & Theoretical & Empirical \\
  \midrule
  1024 & 1.403x & 1.141x & 1.443x & 1.233x \\
  2048 & 1.341x & 1.214x & 1.403x & 1.252x \\
  3072 & 1.296x & 1.203x & 1.370x & 1.238x \\
  4096 & 1.261x & 1.203x & 1.341x & 1.236x \\
  5120 & 1.234x & 1.202x & 1.317x & 1.228x \\
  6144 & 1.212x & 1.192x & 1.296x & 1.215x \\
  7168 & 1.194x & 1.176x & 1.278x & 1.210x \\
  8192 & 1.178x & 1.175x & 1.261x & 1.185x \\
  \bottomrule
  \end{tabular}}
  \end{table}

Two trends are apparent from Table~\ref{tab:flop-speedup-comparison}.
\begin{enumerate}[label=(\roman*),nosep]
    \item The speedup decays with sequence length in both stages. Under this model, the prefill theoretical speedup drops from \(1.403\times\) at \(l=1024\) to \(1.178\times\) at \(l=8192\), while the decode theoretical speedup drops from \(1.443\times\) to \(1.261\times\) over the same range. The empirical results broadly match these theoretical predictions: in both stages, the measured speedups exhibit the same monotonic decay with sequence length, while remaining consistently below the theoretical values due to implementation overheads and computational costs not captured by the FLOP analysis.
    \item The decode speedup is consistently higher than the prefill speedup at the same length. For a fixed \(l\), the unchanged attention cost in decode is smaller than that in prefill, so the reduction in MoE computation accounts for a larger fraction of the total FLOPs and translates into a larger overall speedup. This ordering is consistently reflected in the empirical results across all evaluated lengths.
\end{enumerate}

\end{document}